\documentclass[11pt]{article}

\usepackage[final]{acl}

\usepackage{times}
\usepackage{latexsym}

\usepackage[T1]{fontenc}
\usepackage[utf8]{inputenc}

\usepackage{amsmath,amsfonts,bm}
\usepackage{booktabs}    % for \toprule, \midrule, \bottomrule
\usepackage{amssymb}     % for \checkmark
\usepackage{pifont}      % for extra symbols if needed
\usepackage{natbib}      % for citations
\usepackage{graphicx}    % (optional) if you want to include figures
\usepackage{array}       % better table alignment
\usepackage{caption}     % better table captions
\usepackage{booktabs}    % for \toprule, \midrule, \bottomrule
\usepackage{multirow}    % for \multirow
\usepackage{array}       % for better tables
\usepackage{caption}     % for captions
\usepackage{natbib}      % for references
\usepackage{pifont}
\usepackage{url}
\usepackage{xurl}
\usepackage{hyperref}
\usepackage{makecell}
\usepackage{tabularx}
\usepackage{tablefootnote}
\usepackage{wasysym}     
\usepackage{colortbl}
\newcommand{\cmark}{\textcolor{green!60!black}{\ding{51}}}  % 绿色 ✔
\newcommand{\xmark}{\textcolor{red}{\ding{55}}} % 红色 ✘

\usepackage[table]{xcolor}
\usepackage{ragged2e}
\newcolumntype{Y}{>{\RaggedRight\arraybackslash}X} 
\newcolumntype{C}{>{\Centering\arraybackslash}X}  
 
\definecolor{GrayBG}{gray}{0.95}

\newcolumntype{L}{>{\raggedright\arraybackslash}X}
\def\eqref#1{equation~\ref{#1}}
\def\1{\bm{1}}

\DeclareMathAlphabet{\mathsfit}{\encodingdefault}{\sfdefault}{m}{sl}
\SetMathAlphabet{\mathsfit}{bold}{\encodingdefault}{\sfdefault}{bx}{n}

\usepackage{microtype}
\usepackage{float}
\usepackage{threeparttable}
\AtBeginDocument{\captionsetup{hypcap=true}}
\usepackage{url}
\usepackage{fontawesome}
\usepackage[skins]{tcolorbox}
\tcbuselibrary{listings,breakable}
\usepackage{pifont}
\usepackage{amsthm}
\usepackage{alltt}
\usepackage{booktabs}
\usepackage{siunitx}
\usepackage{cleveref}
\usepackage{kantlipsum}
\usepackage{setspace}
\usepackage{tablefootnote}
\usepackage{amssymb}
\usepackage{pifont} 

\usepackage{xspace}

\newcommand{\hscore}{\textit{$\mathcal{H}$}-Score}

\definecolor{bestbase}{HTML}{009E73}   % Okabe-Ito bluish 
\definecolor{secondbase}{HTML}{56B4E9} % Okabe-Ito sky blue
\colorlet{rank1}{bestbase!16!white}    % Best  (softer)
\colorlet{rank2}{secondbase!16!white}  % Second(softer)
\usepackage{xcolor}

\definecolor{aclblue}{RGB}{64, 112, 175}
\definecolor{aclpaleblue}{RGB}{235, 242, 250} 
\newtcolorbox{PromptBox}[1][]{
  title=Prompt,
  enhanced,
  breakable,
  colback=blue!5,
  colframe=blue!40,  
  coltitle=black,
  fonttitle=\bfseries,
  listing only,
  listing options={basicstyle=\ttfamily\footnotesize, breaklines=true},
  #1
}

\usepackage{inconsolata}

\usepackage{graphicx}
\usepackage{listings}
\usepackage{xcolor} % Optional: for customizing colors
\definecolor{lightergray}{rgb}{0.95, 0.95, 0.95}
\usepackage{tcolorbox}
\usepackage{enumitem}
\usepackage{wrapfig}
\usepackage{longtable}
\definecolor{orange}{RGB}{255,198,140}
\usepackage{pifont}
\usepackage{multirow}
\usepackage{needspace}

\definecolor{aclbluer}{RGB}{52, 73, 94}
\definecolor{aclback}{RGB}{240, 248, 255} 
\definecolor{tblheader}{RGB}{220, 230, 241}

\definecolor{darkgreen}{RGB}{0,100,0}
\definecolor{BoxBlue}{HTML}{1F6FEB}
\definecolor{BoxBlueLight}{HTML}{EAF2FF}

\newtcolorbox{ReasonBox}[1]{
  enhanced,
  breakable,
  sharp corners,
  boxrule=0.8pt,
  colframe=BoxBlue,
  colback=BoxBlueLight,
  coltitle=white,
  fonttitle=\bfseries,
  title=#1,
  attach boxed title to top left={yshift=-2mm, xshift=4mm},
  boxed title style={
    colback=BoxBlue,
    colframe=BoxBlue,
    sharp corners,
    boxrule=0pt,
    left=6pt,right=6pt,top=4pt,bottom=4pt
  },
  left=10pt,right=10pt,top=10pt,bottom=10pt
}
\newcommand{\dR}[1]{{\tiny\textcolor{red}{(#1)}}}              
\newcommand{\dG}[1]{{\tiny\textcolor{green!60!black}{(#1)}}}

\newcommand{\cellD}[2]{\makecell[c]{#1\\[-1pt]#2}}
\newcommand{\cellB}[1]{\makecell[c]{#1\\[-1pt]{\tiny\phantom{(+00.00)}}}}

\title{
    \includegraphics[height=2em]{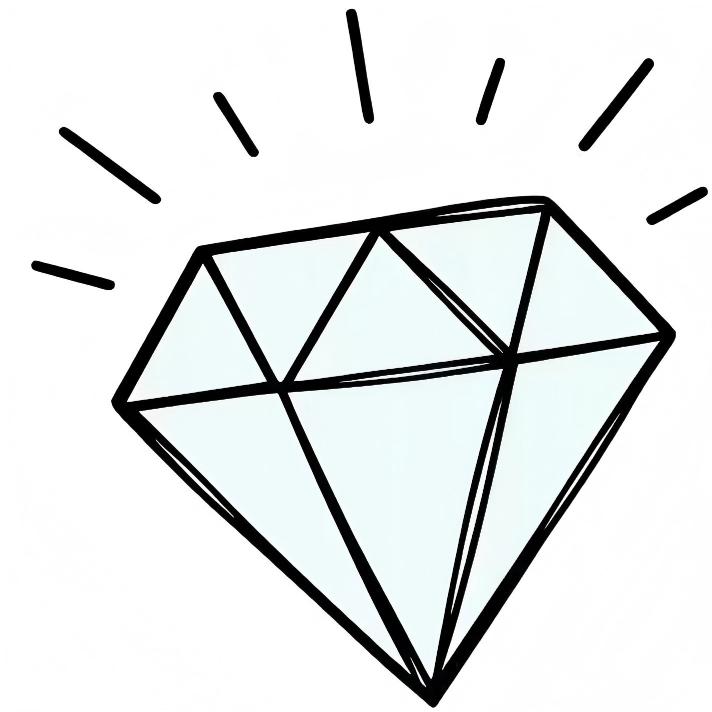}
    \space
    PRISM: Probing Reasoning, Instruction, and Source Memory in LLM Hallucinations
}

\author{
 \textbf{Yuhe Wu\textsuperscript{1}},
 \textbf{Guangyu Wang\textsuperscript{2,3}},
 \textbf{Yuran Chen\textsuperscript{3}},
 \textbf{Jiatong Zhang\textsuperscript{3}},
 \textbf{Yutong Zhang\textsuperscript{3}},\\
 \textbf{Yujie Chen\textsuperscript{4}},
 \textbf{Jiaming Shang\textsuperscript{5}},
 \textbf{Guang Zhang\textsuperscript{1}\thanks{Corresponding authors.\protect\hypertarget{corrauthor}{}}},
 \textbf{Zhuang Liu\textsuperscript{3}\hyperlink{corrauthor}{\textsuperscript{*}}}
\\
\\
 \textsuperscript{1}HKUST(GZ)\quad
 \textsuperscript{2}NYUSH\quad
 \textsuperscript{3}DUFE\\
 \textsuperscript{4}CUHK(SZ)\quad
 \textsuperscript{5}CUFE\\
  \small{
   \faEnvelopeO~: \href{liuzhuang@dufe.edu.cn}{guangzhang@hkust-gz.edu.cn, liuzhuang@dufe.edu.cn} \quad \faGlobe: \href{ https://acl-prism.cc}{https://acl-prism.cc}
 }
}

\begin{document}
\maketitle

\begin{abstract}
As large language models (LLMs) evolve from conversational assistants into agents capable of handling complex tasks, they are increasingly deployed in high-risk domains. 
However, existing benchmarks largely rely on mixed queries and posterior evaluation, output-level scoring, which quantifies hallucination severity but offers limited insight into \emph{where} and \emph{why} hallucinations arise in the generation pipeline. 
We therefore reformulate hallucination evaluation as a diagnostic problem and propose \textbf{\textit{PRISM}}, a controlled benchmark 
that disentangles hallucinations into four dimensions: knowledge missing, knowledge errors, reasoning errors, and instruction-following errors, grounded in three stages of generation (memory, instruction, and reasoning). 
\textbf{\textit{PRISM}} contains 9{,}448 instances across 65 tasks and supports fine-grained, stage-aware diagnostic evaluation. Evaluating 24 mainstream open-source and proprietary LLMs, we uncover consistent trade-offs across instruction following, memory retrieval, and logical reasoning, showing that mitigation strategies often improve specific dimensions at the expense of others.
We hope \textbf{\textit{PRISM}} provides a framework for understanding the specific mechanisms behind LLMs hallucinations, ultimately accelerating the development of trustworthy large language models.
\end{abstract}

\section{Introduction}
LLMs have become capable of handling complex tasks \citep{wang2024survey, zhang-etal-2025-planning,ijocLIU2025mitigating, xi2023rise}, facilitating their application in high-risk domains such as medical diagnosis \citep{singhal2023large, thirunavukkarasu2023large}, legal consulting \citep{guha2024legalbench,cui2023chatlaw}, and scientific discovery \citep{boiko2023autonomous,bran2023chemcrow}.
While current models perform well on general benchmarks \citep{hendrycks2021measuring, achiam2023gpt}, they frequently generate factually inconsistent content \citep{alansari2025largelanguagemodelshallucination} when encountering outdated concepts \citep{kandpal2023large, mallen2023not}, dynamic information \citep{kasai2024realtimeqawhatsanswer, vu2023freshllms}, or complex reasoning and instruction constraints \citep{dziri2024faith, lanham2023measuring, heyman2025reasoninglargelanguagemodel}. 
Such unfaithfulness not only erodes user trust but also constitutes potential safety hazards in critical decision-making scenarios \citep{thirunavukkarasu2023large, zhang2023siren}.  Consequently, the evaluation of hallucinations has emerged as a fundamental challenge that the research community needs to overcome.

Despite the growing interest in quantifying hallucinations \citep{lin2022truthfulqa, li2023halueval}, existing benchmarks have clear limitations in answering the fundamental question of why models fail.
% While representing a significant leap from general capability evaluation to safety assessment, these benchmarks suffer from two limitations as follows:

\begin{figure*}[t]
    \centering
    \includegraphics[width=1\linewidth]{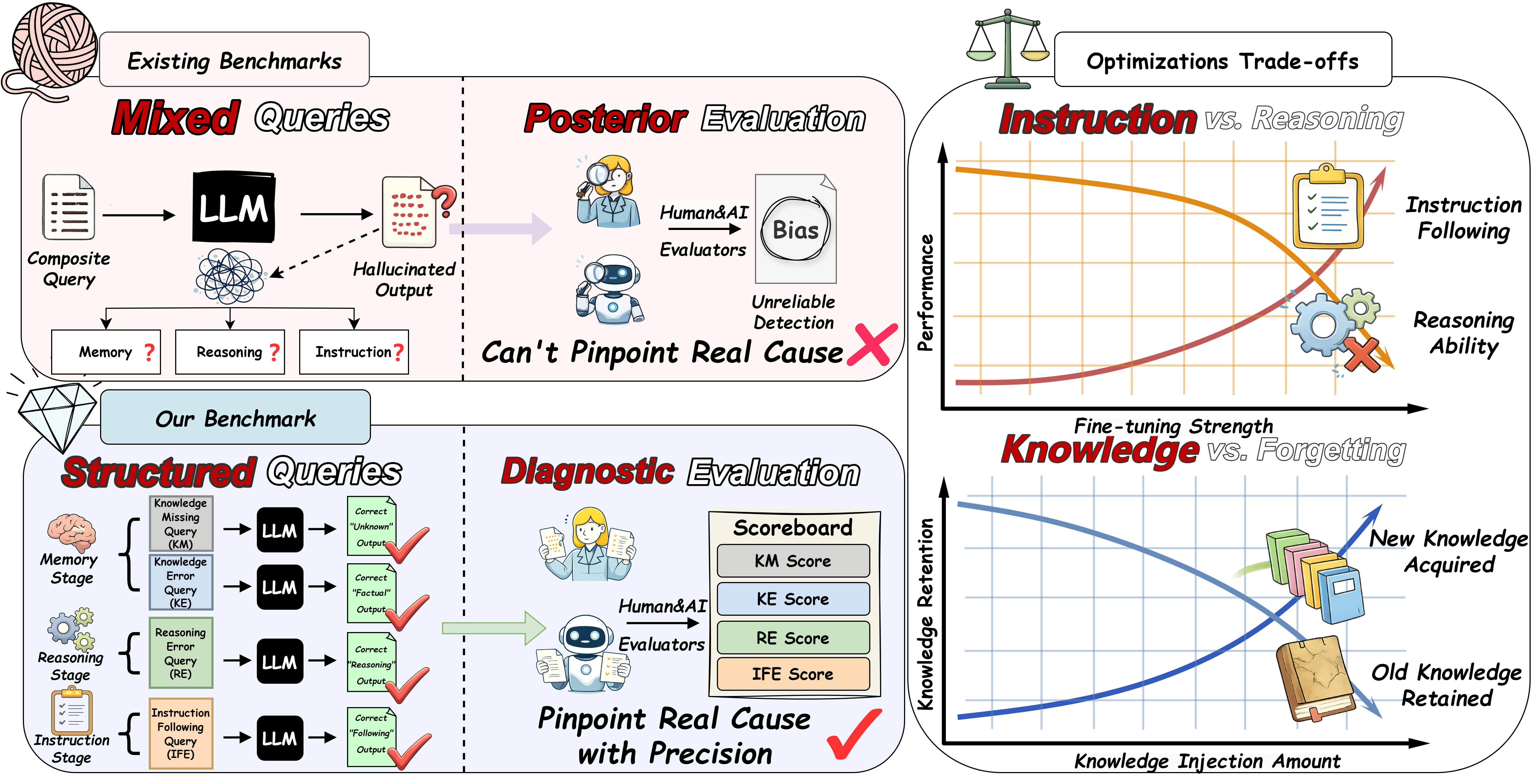}
    \caption{Overview of the \textbf{\textit{PRISM}} framework and optimization trade-offs. The left panel contrasts the mixed query design of existing benchmarks with our structured approach that isolates cognitive stages to pinpoint failure dimensions like KE, KM, RE, and IFE. The right panel illustrates performance trade-offs where enhancing instruction following compromises reasoning ability and knowledge injection leads to the forgetting of retained information.}
    \label{fig:1}
\end{figure*}

First, current benchmarks often mix different queries, which prevents us from testing skills in isolation. As shown in Figure~\ref{fig:1} (Left Top), benchmarks like TruthfulQA \citep{lin2022truthfulqa}, HaluEval \citep{li2023halueval}, and FreshQA \citep{vu2023freshllms} typically use mixed queries. When a model fails on these, the reason is ambiguous: did it fail to retrieve the right data, make a logical error, or simply ignore the instructions? Second, most evaluations focus only on the final output. Even detailed methods like FActScore \citep{min2023factscore}, and HALoGEN \citep{ravichander2025halogen} depend on posterior evaluation after generation. Relying on such outcome assessments introduces unavoidable bias from both human and model evaluators. Furthermore, this approach fails to test specific inputs to isolate the error. Without knowing exactly where the process broke, fixing the model is much harder. An abstract comparison is shown in Table \ref{tab:hallucination-benchmark}.

As illustrated in Figure \ref{fig:1} (Right), improvement strategies for different mechanisms often involve inherent trade-offs: for instance, strong instruction fine-tuning to fix formatting errors can accidentally hurt rigorous reasoning capabilities \citep{ouyang2022training, peng2023instructiontuninggpt4}, while indiscriminate knowledge injection may cause catastrophic forgetting \citep{zhai2023investigatingcatastrophicforgettingmultimodal}. Consequently, to achieve reasonable optimization, we must address a fundamental question:
\begin{tcolorbox}[
    enhanced,
    frame hidden,
    colback=aclpaleblue,
    boxrule=0pt,
    borderline west={2pt}{0pt}{aclblue},
    arc=1pt,
    left=4pt, right=4pt, top=4pt, bottom=4pt, 
]
    \linespread{1.1}\selectfont
    \textcolor{black!90}{
        \textcolor{aclblue}{\faCrosshairs}\hspace{4pt} 
        \textit{\textbf{How can we establish a framework to pinpoint failures in memory, reasoning, and instruction following, thereby guiding optimization for hallucination mitigation?}}
    }
\end{tcolorbox}

To address the aforementioned challenges and establish a trustworthy diagnostic framework, we propose \textit{\textbf{PRISM}}, an evaluation benchmark grounded in the interactive pipeline of LLMs.
Based on the three generation stages of instruction following, memory retrieval, and reasoning, we exactly categorize hallucination phenomena into four independent failure dimensions:

\begin{itemize}
    \setlength\itemsep{0em}
    \item \textbf{Knowledge Error (KE):} The model's parametric knowledge stores incorrect or outdated information.
    \item \textbf{Knowledge Missing (KM):} The model's parametric knowledge lacks the correct information required to answer the question.
    \item \textbf{Reasoning Error (RE):} The model possesses the necessary facts but fails to combine them through logic or reasoning.
    \item \textbf{Instruction Following Error (IFE):} The model possesses correct knowledge and reasoning capabilities, but its output violates explicit constraints provided by the user.
\end{itemize}

This design enables us to locate specific weaknesses of the model in the generation stages. Our main contributions are summarized as follows:

\begin{itemize}
    \setlength\itemsep{0em}
\item We propose a cognitive pipeline failure framework that defines hallucinations as dimensions in KE, KM, RE, and IFE. We then build \textit{\textbf{PRISM}}, a benchmark of 9.448 samples that isolates these factors and pinpoints model weaknesses for reproducible analysis.

\item We conducted a comprehensive evaluation of 24 proprietary and open-source LLMs across 4 dimensions and 65 sub-tasks to assess the causes of hallucinations in different model types and encourage the training of hallucination-specific LLMs.

\item Building on \textit{\textbf{PRISM}}, we examined the performance trade-offs of common hallucination mitigation strategies. Furthermore, we constructed a toy dataset to support case-based empirical studies, allowing us to reveal the internal mechanisms of LLMs during KE and KM memory-based issues and analyze the relationship between IFE and LLMs efficiency. These findings guide the design of balanced mitigation strategies.

\end{itemize}

\section{Benchmark Construction}
To achieve precise attribution of hallucination mechanisms, our data construction follows the principle of orthogonality, meaning that each data subset aims to independently test a single failure mode to the greatest extent possible.

\subsection{Data Source}

To achieve attribution of hallucination dimensions, \textbf{\textit{PRISM}}'s data construction strictly follows the orthogonality principle: each subset is designed to test a single failure mode. As shown in Table~\ref{tab:hallucination-benchmark}, existing benchmarks suffer from limited evaluation scope and a lack of variable control, making metrics incapable of revealing the causes of errors. Consequently, we construct a corpus that strictly partitions data into parametric knowledge-dependent and Reasoning and instruction-dependent categories, enabling the isolated probing of specific hallucination modes. Further detailed source lists are available in Appendix~\ref{source}.

\begin{table*}[t]
\centering
\small
\resizebox{\textwidth}{!}{
\begin{tabular}{l c lc c c c }
\toprule
\multirow{2}{*}{\textbf{Benchmark}} &
\multicolumn{4}{c}{\textbf{Evaluation Scope}}&
\multicolumn{2}{c}{\textbf{Methodological Design}}\\

\cmidrule(lr){2-5}
\cmidrule(lr){6-7}

& \textbf{KE}&  \textbf{KM}&\textbf{RE}& \textbf{IFE}& \textbf{Variable Control} & \textbf{Diag. Mode}\\

\midrule
TruthfulQA \citep{lin2022truthfulqa} & \cmark &  \xmark
&\xmark & \xmark
& \xmark & \xmark
\\

HaluEval \citep{li2023halueval} & \cmark &  \cmark
&\cmark & \xmark
& \xmark & \xmark
\\

FActScore \citep{min2023factscore} & \cmark &  \cmark
&\xmark & \xmark
& \xmark & \xmark
\\

FELM \citep{chen2023felm} & \cmark &  \xmark
&\cmark & \xmark
& \xmark & \xmark
\\

FreshQA \citep{vu2023freshllms} & \cmark &  \cmark
&\xmark & \xmark
& \cmark & \xmark
\\

FollowBench \citep{jiang2024followbench} & \xmark &  \xmark
&\cmark & \cmark
& \xmark & \xmark
\\

HALoGEN \citep{ravichander2025halogen} & \cmark &  \cmark
&\cmark& \xmark
& \xmark & \xmark
\\

HalluLens \citep{bang2025hallulens} & \cmark &  \cmark
&\cmark & \xmark
& \xmark & \xmark
\\

\midrule
\rowcolor{aclblue!10}
\textbf{\textit{PRISM (Ours)}} & \cmark &  \cmark
&\cmark & \cmark
& \cmark~& \cmark
\\

\bottomrule
\end{tabular}}

%\caption*{\footnotesize \textbf{\textit{Note:}} \textbf{Variable Control} denotes whether a benchmark enforces orthogonal isolation of causal factors, ensuring attribution correctness by preventing confounding effects when decomposing hallucination sources.}

\caption{Comparison of hallucination evaluation benchmarks.
\textbf{\textit{PRISM}} uniquely achieves \emph{comprehensive evaluation scope}, \emph{causative explainability}, and \emph{decoupled probing-based diagnosis}.}

\label{tab:hallucination-benchmark}
\end{table*}

\paragraph{Sources for Parametric Knowledge Tasks.}
This category aims to define the accuracy and boundaries of the model's internal memory by comparing it against external objective facts. We collected two types of raw corpus:

\begin{itemize}[leftmargin=*,labelsep=5pt]

\item \textbf{Factual Data:} To ensure the factual consistency of our evaluation standards, we selected \textit{Wikipedia} and \textit{Baidu Baike} as the primary sources for KE tasks. Compared to unfiltered web texts, these corpora have lower noise. We focused on collecting long-tail and ambiguous entries that are not in the LLM's source memory base, used to test the model's memory coverage and the ability to distinguish specific entities.

\item \textbf{Out-of-Distribution (OOD) Data:} To evaluate the model's ability to identify unknown information, we first established a temporal news corpus by collecting news reports and paper abstracts published by CNN, Reuters, and arXiv between March 2025 and November 2025. As this period postdates the training cutoff of most baseline models, these materials constitute a test source for future information. 
Second, we constructed a fictional entity corpus. Rather than being collected from the real world, this content consists of generated counterfactual descriptions created by setting specific attributes. Finally, to cover information that is real but non-public, we introduced private domain data.

\end{itemize}

\paragraph{Sources for Reasoning and Instruction Tasks.}
This type of data is designed to reduce the model's reliance on parametric knowledge and to focus on evaluating its ability to perform logical reasoning and execute rules under a given context. We built two corpora to keep the focus on the reasoning process rather than memory retrieval.

\begin{itemize}[leftmargin=*,labelsep=5pt]

\item \textbf{Self-Contained Reasoning Data:} To ensure the reasoning process is isolated from external knowledge noise, we prioritized task types where the solution premises are strictly embedded within the input context. In addition to introducing competition problems, such as the IMO, to cover formal logic and mathematical proofs, we also incorporated code generation tasks. Given their deterministic execution logic, they provide the most ideal ambiguity-free reasoning environment for the model.

\item \textbf{Complex Instruction Data:} Beyond covering everyday basic instructions, we specifically construct a high-constraint corpus, and we use automated templates to generate adversarial synthetic data. This corpus simulates real scenarios where instruction violations happen due to competition for attention resources under multi-dimensional stacked constraints, including negative semantics that forbid specific words, format locking that requires strict JSON output, and various limits on length and language.

\end{itemize}

\subsection{Construction Pipeline}

\begin{figure*}[t]
    \centering
    \includegraphics[width=0.95\linewidth]{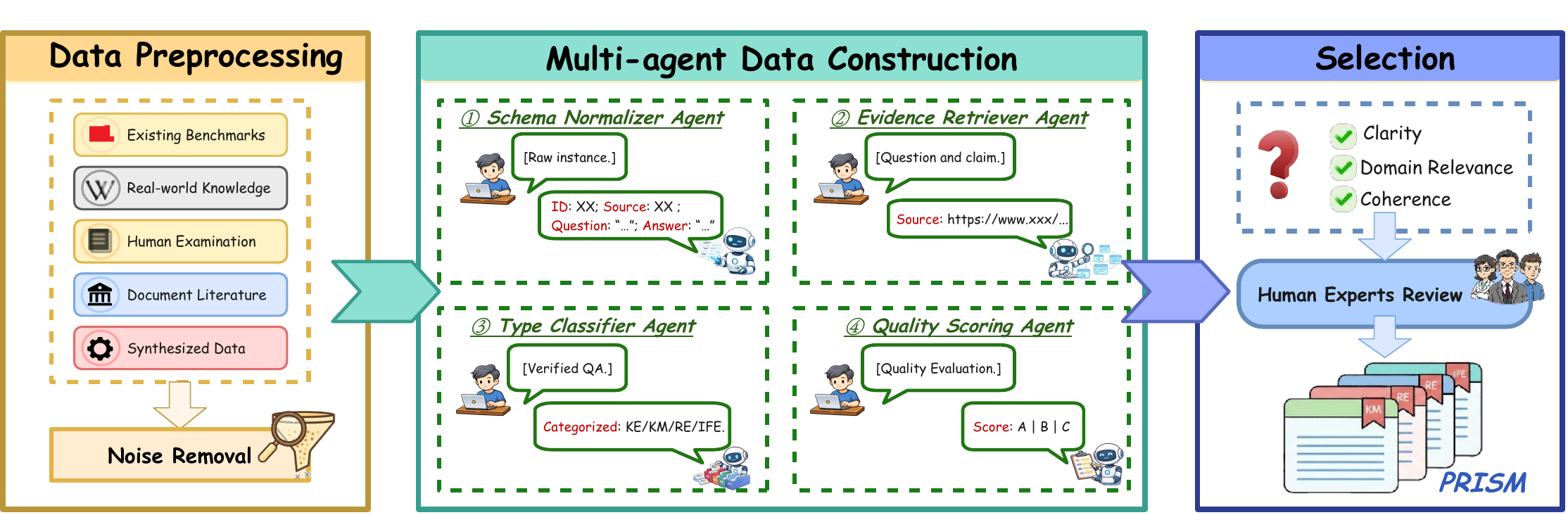}
    \caption{The Three-phase Pipeline of \textbf{\textit{PRISM}} Benchmark Construction}
    \label{11111}
\end{figure*}

To construct \textbf{\textit{PRISM}}, we design a three-stage pipeline as illustrated in Figure ~\ref{11111}. 

% In the data collection stage, a corpus is collected from authoritative sources and cleaned via noise removal. 
% In the multi-agent data construction stage, We present a four-agent methodology to construct data. The schema normalizer agent standardizes raw inputs into a unified format, while the evidence retriever agent verifies each data is accompanied by a coherent proof or derivation. Building on the verified corpus, the type classifier agent assigns each instance to a specific failure type and the quality scoring agent subsequently evaluates instances along multiple dimensions, assigning coarse-grained quality labels for filtering. 
% In the human selection stage, domain experts further review the remaining instances with respect to category alignment, result alignment, and semantic coherence. Only instances that satisfy all criteria are retained, resulting in a high-quality PRISM benchmark. Detailed construction procedures are provided in Appendix ~\ref{pipline}.

\paragraph{Data Collection.} In this initial stage, a corpus is collected from authoritative sources and cleaned via noise removal.
    
\paragraph{Multi-agent Data Construction.} Next, we adopt a multi-agent framework to construct data: (\romannumeral1) schema normalizer agent; (\romannumeral2) evidence retriever agent; (\romannumeral3) type classifier agent; (\romannumeral4) quality scoring agent, enabling each agent to focus on a specific step for improving the results, thereby enhancing both the efficiency and quality of the data.

\paragraph{Human Selection.} 
Domain experts select instances for clarity, relevance, and coherence to curate the \textbf{\textit{PRISM}}. Detailed construction procedures are provided in Appendix~\ref{pipline}.

\subsection{Data Statistics}

\label{sec:statistics}

\textbf{\textit{PRISM}} contains a total of 9,448 evaluation instances, covering 4 failure dimensions and 65 specific sub-tasks. Among them, 2,995 are RE samples (31.7\%), 2,442 are IFE samples (25.8\%), 2,078 are KM samples (22.0\%), and 1,933 are KE samples (20.5\%).

\begin{figure*}[t]
    \centering
    \includegraphics[width=\linewidth]{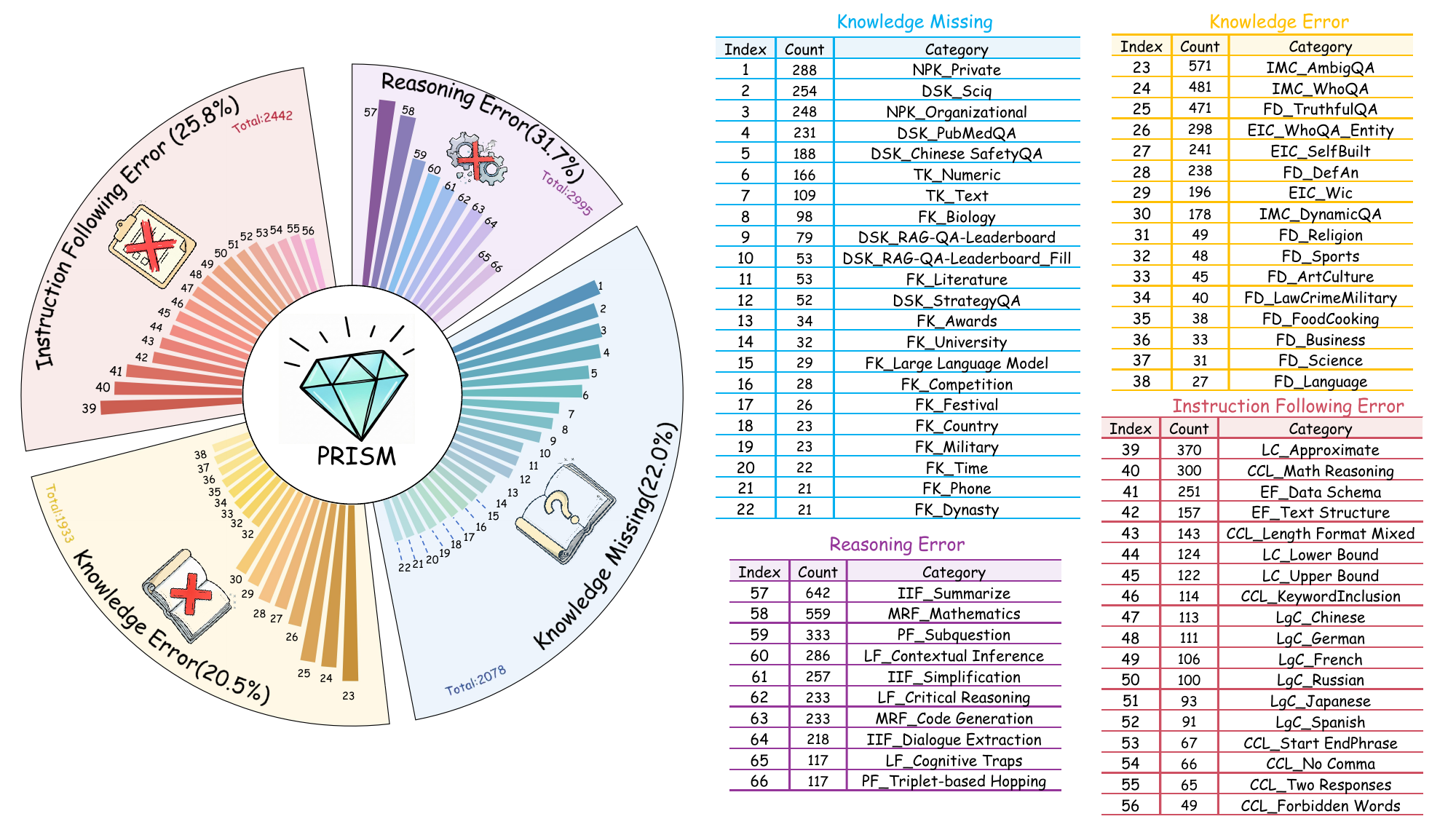}
    \caption{The hierarchical distribution of \textbf{\textit{PRISM}}. The inner circle represents the four primary failure dimensions, while the outer ring details 65 sub-tasks. For consistency, we define the abbreviations as follows: DSK = Domain-Specific Knowledge, FK = Fictional Knowledge, TK = Timely Knowledge, NPK = Non-Public Knowledge, FD = Factual Distortion, IMC = Intra-Memory Conflict, EIC = Entity-Identity Confusion, LF = Logical Fallacy, PF = Procedural Failure, IIF = Information Integration Failure, MRF = Mathematical Reasoning Failure, EF = Explicit Format, LC = Length Constraints, LgC = Language Constraints, and CCL = Complex \& Cognitive Load.}
    \label{fig:statistics}
\end{figure*}

The distribution of the data is illustrated in Figure~\ref{fig:statistics}. The left side of the figure displays a sunburst chart where the inner circle represents the four primary failure dimensions, and the outer ring corresponds to sub-task indices ranging from 1 to 65. The right side of the figure lists the detailed mapping for each index, specifying the category name and the exact sample count for each sub-task.

\section{Experiments}
\textbf{\textit{PRISM}} is designed to evaluate the reliability of LLMs and provide guidance for model optimization. To meet these objectives, we structure our experiments around three research questions that establish performance baselines, explain underlying causes, and identify pathways for improvement:
\begin{itemize}[leftmargin=*,labelsep=5pt]
    \item \textbf{RQ1:} What is the overall performance of LLMs on \textbf{\textit{PRISM}}, and how do different error types vary across model families and scales?
    \item \textbf{RQ2:} How effective are common mitigation strategies, and which methods best address specific types of hallucinations?
    \item \textbf{RQ3:} Why do different types of hallucinations show consistent patterns in the internal representations of LLMs?
    % wgy & wyh
    % 换顺序，RQ3: Toydataset attention map https://arxiv.org/pdf/2509.20942
    %  case:
    % 缺失 & 错误：加扰动-错误文本 attention map 10  10 
    %  效率分析： iFE: 拒答情况下激活神经元数量 & 深度 & FLOPS 10
    %RQ2 Qwen-3-4B, Llama-3.1-8B(Base.Instruct,SFT,ICL)
\end{itemize}
To provide a solid foundation for investigating these questions, we begin by outlining the experimental setup, focusing on model selection and evaluation protocols.

\subsection{Evaluation Setup}
\paragraph{Model Selection.} We evaluate 24 representative LLMs under the few-shot setting.
To ensure comprehensive coverage, the evaluated LLM series includes both open-source and proprietary models, such as GPT~\cite{gpt_2024}, Gemini~\cite{gemini_2025}, Llama~\cite{llama_2025}, Claude~\cite{claude_2025}, DeepSeek~\cite{deepseek_2025}, GLM~\cite{glm_2024}, Qwen~\cite{qwen_2024}, and Grok~\cite{grok_2025}.

\paragraph{Evaluation Metrics.}

We employ distinct metrics for each subtask to enable a hallucination comparison.

\begin{itemize}
    \item \textbf{Accuracy:} For closed-ended tasks, we employ standard Accuracy:
    \[
        \mathrm{Acc}=\frac{1}{N}\sum_{i=1}^{N}\mathbb{I}\!\left[\hat{y}_i=y_i\right]
    \]

    \item \textbf{LLM-Eval:} For open-ended tasks, we adopt a LLM evaluator following \textsc{LLM-Eval} \citep{lin2023llmeval,zheng2023judging}, which produces a scalar score $s\in[0,5]$. The prompts and the consistency evaluation between the LLM-as-a-judge and human annotations are provided in Appendix~\ref{llm_evaluation}.

    \item \textbf{Hallucination Rate:} We first map all task metrics to a unified percentage score $S\in[0,100]$:
    \[
    S =
    \begin{cases}
        100\cdot \mathrm{Acc}, & \text{for closed-ended tasks},\\[2pt]
        100\cdot \dfrac{s}{5}, & \text{for open-ended tasks},
    \end{cases}
    \]
    and define the hallucination rate as its complement:
    \[
    \mathcal{H} = 100 - S.
    \]

    \item \textbf{\hscore{}:} Let $\mathcal{H}_d$ denote the macro-averaged hallucination rate for each dimension
    $d\in D=\{\mathrm{KE},\mathrm{KM},\mathrm{RE},\mathrm{IFE}\}$.
    We define
    \[
        \hscore{}=\frac{1}{4}\sum_{d\in D}\mathcal{H}_d.
    \]
\end{itemize}

\paragraph{Sampling Parameters.}
To ensure fair and comparable evaluations across diverse models, we control generation randomness using temperature and top $p$ sampling \citep{holtzman2019curious,openai_api_top_p}, exploring temperature values in \{0, 0.2, 0.4, 0.6, 0.8\} and top-p values in \{0.6, 0.8, 0.9, 0.95\}. For closed-ended tasks, we conducted the grid search on the DeepSeek-R1-Distill-32B , identifying temperature = 0.8 and top-p = 0.8 as optimal for the KM, KE, and IFE subsets, while adopting temperature = 0.4 and top-p = 0.95 for RE to enhance reasoning stability. For open-ended tasks, we performed the grid search on GPT-4o, yielding temperature = 0.8 and top-p = 0.8 for peak average performance. The full grid results are reported in Appendix~\ref{app:decoding_grid}.

\subsection{Evaluation Results (RQ1)}
\begin{figure}[t]
\centering
\includegraphics[width=0.85\linewidth]{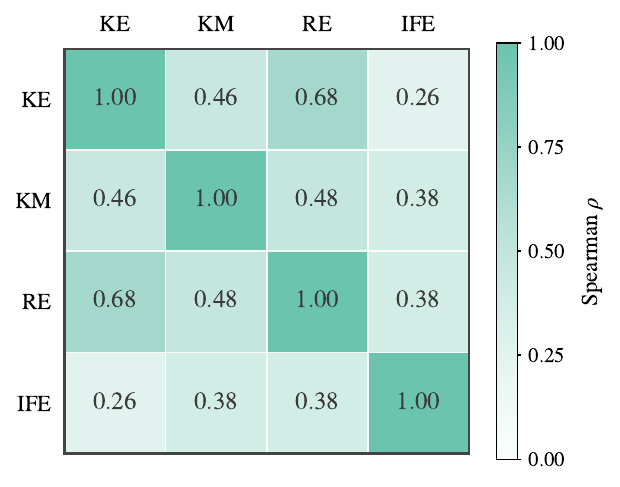}
\caption{Spearman Correlation of Model rankings
across 4 Dimensions}
\label{fig:dim-corr}
\end{figure}

Table~\ref{tab:prism-main-complete} comprehensively reports the hallucination rates across four core evaluation dimensions along with the composite \hscore\ metrics. Claude-Opus-4.5, Gemini-3-Pro, and Gemini-3-Flash achieve the top three rankings with the lowest \hscore s (13.90\%, 14.29\%, and 15.55\%, respectively), indicating more robust and balanced performance across diverse error types. In terms of dimensional distribution, RE and KE exhibit significant performance gaps among models. In contrast, error rates in the KM dimension remain relatively low for most models, rendering this dimension less of a dominant factor in determining the final ranking. The analysis in Figure~\ref{fig:dim-corr} further reveals only partial consistency across dimensions: KE shows the strongest correlation with RE and the weakest with IFE. More detailed experimental results are provided in Appendix~\ref{result_table}.

In the comparison of model types, open-source models have gradually approached proprietary models in the KM dimension but still lag significantly in RE and IFE, particularly on complex tasks requiring the integration of multiple clues and consistent reasoning. Furthermore, experiments indicate that adopting step-by-step explanation strategies does not yield consistent performance gains. This is likely because the hallucination phenomenon is complex, encompassing factual errors, inconsistent reasoning logic, misuse of evidence, and deviation from task requirements during complex instruction following \citep{ji2023survey, huang2023survey}. Additionally, the generated step-by-step explanations may appear plausible on the surface but do not faithfully reflect the internal information upon which the model actually relies during generation \citep{turpin_language_2023}.

\begin{table*}[t]
\centering
\small
\setlength{\tabcolsep}{5.5pt}
\renewcommand{\arraystretch}{1.07}
\begin{tabular}{l l c c cccc | c}
    \toprule
    \multirow{2}{*}{\textbf{Type}} &
    \multirow{2}{*}{\textbf{Model}} &
    \multirow{2}{*}{\textbf{Size}} &
    \multirow{2}{*}{\textbf{Think}} &
    \textbf{KE} & \textbf{KM} & \textbf{RE} & \textbf{IFE} & \textbf{\hscore} \\
    \cmidrule(lr){5-9}
      & & & &
      \multicolumn{5}{c}{{(Lower the Better)}} \\
    \midrule
    % -------- Proprietary LLMs --------
    \multirow{12}{*}{\textit{\textbf{Proprietary LLMs}}}
      & GPT-5.2 & - & \cmark & 17.83\% & \cellcolor{rank1}5.57\% & 24.67\% & 15.35\% & 15.85\% \\
      & GPT-5.1 & - & \cmark & 25.83\% & 11.46\% & 26.01\% & 25.62\% & 22.23\% \\
      & GPT-4o-20241120 & - & \xmark & 19.36\% & 8.90\% & 29.72\% & 16.78\% & 18.69\% \\
      & Gemini-3-Pro & - & \cmark & 16.31\% & 12.80\% & \cellcolor{rank1}17.19\% & \cellcolor{rank1}10.85\% & \cellcolor{rank2}14.29\% \\
      & Gemini-3-Flash & - & \cmark & 16.48\% & 8.12\% & 21.59\% & 16.00\% & 15.55\% \\
      & Gemini-2.5-Pro & - & \cmark & 25.33\% & 8.99\% & 21.63\% & \cellcolor{rank2}13.84\% & 17.45\% \\
      & Gemini-2.5-Flash & - & \cmark & 19.29\% & 8.74\% & 31.58\% & 15.51\% & 18.78\% \\
      & Claude-Opus-4.5 & - & \cmark & \cellcolor{rank1}13.80\% & \cellcolor{rank2}6.35\% & \cellcolor{rank2}19.68\% & 15.77\% & \cellcolor{rank1}13.90\% \\
      & Claude-Sonnet-4.5 & - & \cmark & \cellcolor{rank2}16.11\% & 7.03\% & 23.53\% & 16.87\% & 15.89\% \\
      & Claude-Haiku-4.5 & - & \cmark & 18.13\% & 7.24\% & 25.59\% & 18.15\% & 17.28\% \\
      & Grok-4.1 & - & \xmark & 19.20\% & 17.03\% & 34.94\% & 20.04\% & 22.80\% \\
      & Grok-4-0709 & - & \xmark & 18.35\% & 14.57\% & 30.97\% & 15.74\% & 19.91\% \\
    \midrule
    % -------- Open-source LLMs --------
    \multirow{12}{*}{\textit{\textbf{Open-source LLMs}}}
      & DeepSeek-V3.2 & 685B & \xmark & \cellcolor{rank2}17.99\% & 7.94\% & 28.31\% & 14.31\% & \cellcolor{rank2}17.14\% \\
      & DeepSeek-R1 & 671B & \cmark & 18.87\% & 10.13\% & 28.04\% & \cellcolor{rank1}11.97\% & 17.25\% \\
      & DeepSeek-R1-Distill-32B & 32B & \cmark & 18.14\% & 11.27\% & 30.40\% & 18.71\% & 19.63\% \\
      & Qwen3-235B-Instruct & 235B & \cmark & \cellcolor{rank1}17.37\% & \cellcolor{rank2}7.00\% & \cellcolor{rank1}25.89\% & 17.87\% & \cellcolor{rank1}17.03\% \\
      & Qwen2.5-72B-Instruct & 72B & \xmark & 18.56\% & 8.61\% & \cellcolor{rank2}27.68\% & 18.43\% & 18.32\% \\
      & GLM-4.5 & 355B & \cmark & 18.85\% & 10.70\% & 29.34\% & 15.06\% & 18.49\% \\
      & GLM-4 & 32B & \xmark & 26.20\% & 16.17\% & 35.57\% & 25.34\% & 25.82\% \\
      & Llama-4-Scout & 17B & \xmark & 23.67\% & 7.58\% & 32.19\% & 14.17\% & 19.40\% \\
      & Llama-3.3-70B-Instruct & 70B & \xmark & 19.75\% & \cellcolor{rank1}6.24\% & 29.47\% & \cellcolor{rank2}13.13\% & 17.15\% \\
      & Llama-3.1-8B-Instruct & 8B & \xmark & 25.53\% & 14.04\% & 54.50\% & 23.36\% & 29.36\% \\
      & Llama-3-70B-8192 & 70B & \xmark & 20.70\% & 7.43\% & 37.50\% & 15.83\% & 20.37\% \\
      & Llama-3-8B-Instruct & 8B & \xmark & 23.15\% & 18.42\% & 54.02\% & 32.53\% & 32.03\% \\
    \bottomrule
\end{tabular}

\caption{\textbf{Main Results on Hallucination Rates.}
All values are hallucination rates $\mathcal{H}$ (\%) aggregated over 4 dimensions, together with \hscore{}.
Models are highlighted with \colorbox{rank1}{\strut Best} and \colorbox{rank2}{\strut Second Best} within each group.}
\label{tab:prism-main-complete}
\end{table*}

\section{Discussion}
Our experimental results reveal a complex relationship between model parameter scale, reasoning capability, and performance. This section first addresses RQ2 by analyzing how different mitigation strategies affect model capabilities. Building on this analysis, we then turn to RQ3 to investigate the potential causes of different hallucination patterns.

\subsection{Impact of Mitigation Strategies (RQ2)}
To answer RQ2, we examine how hallucination mitigation strategies affect model capability along three dimensions. The first dimension is in-context learning (ICL), where we vary the number of demonstrations to adjust the strength of contextual guidance. The second dimension is instruction tuning, where we introduce Llama-3.1-8B-Instruct to assess behavioral differences induced by general alignment. The third dimension is reasoning tuning, implemented as supervised fine-tuning (SFT) on the reasoning dataset\footnote{Trained on the gsm8k-reasoning dataset. Available at: \url{https://huggingface.co/datasets/thesven/gsm8k-reasoning}}, to study how reasoning enhancement influences different error types. To control variables, the latter two settings are compared under the 1-shot configuration. This design aims to reveal whether optimizations targeting instruction following or reasoning trigger cross-dimensional capability tradeoffs.

Table~\ref{tab:shot-ablation-delta-compact} further shows that none of the three strategies delivers stable improvements across dimensions. Instead, the results form a tradeoff structure. Adjusting shots in ICL barely changes the overall conclusion. The 0-shot setting slightly degrades performance, while the 3-shot setting brings marginal positive changes, suggesting that additional demonstrations mainly aid format rather than addressing root causes such as knowledge gaps. The Instruct model reduces hallucinations more consistently in knowledge conflict and instruction following dimensions, but it introduces a certain level of loss in the reasoning dimension. By contrast, the Reasoning model achieves the strongest improvements in the reasoning dimension, while showing clear capability degradation in knowledge-related and instruction-related dimensions. 
% These trends are further confirmed on Qwen3-VL-32B, which has a completely different architecture and parameter scale, suggesting that the observed tradeoff structure is not model-specific.

\begin{table}[ht]
\centering
\scriptsize
\setlength{\tabcolsep}{3.0pt}
\renewcommand{\arraystretch}{1.08}
\resizebox{0.5\textwidth}{!}{
\begin{tabularx}{\columnwidth}{@{}p{1.6cm} p{1.0cm} *{5}{>{\centering\arraybackslash}X}@{}}
\toprule
\multirow{3}{*}{\textbf{Model}} & \multirow{3}{*}{\textbf{Dataset}} & \textbf{Base} & \multicolumn{2}{c}{\textbf{ICL}} & \textbf{Instruct.} & \textbf{Reason.} \\
\cmidrule(lr){3-3}\cmidrule(lr){4-5}\cmidrule(lr){6-6}\cmidrule(lr){7-7}
 & & \textbf{1-shot} & \textbf{0-shot} & \textbf{3-shot} & \textbf{1-shot} & \textbf{1-shot} \\
\midrule

\multirow{4}{*}{\textbf{Llama3.1-8B}}
& \textbf{KE} &
\cellB{43.15} &
\cellD{43.88}{\dG{+0.73}} &
\cellD{43.04}{\dG{-0.11}} &
\cellD{38.61}{\dR{-4.54}} &
\cellD{67.02}{\dG{+23.87}} \\

& \textbf{KM} &
\cellB{11.95} &
\cellD{13.28}{\dG{+1.33}} &
\cellD{11.79}{\dR{-0.16}} &
\cellD{11.78}{\dR{-0.17}} &
\cellD{29.56}{\dG{+17.61}} \\

& \textbf{IFE} &
\cellB{35.28} &
\cellD{35.53}{\dG{+0.25}} &
\cellD{35.03}{\dR{-0.25}} &
\cellD{33.76}{\dR{-1.52}} &
\cellD{61.87}{\dG{+26.59}} \\

& \textbf{RE\_Math} &
\cellB{94.99} &
\cellD{95.71}{\dG{+0.72}} &
\cellD{94.81}{\dR{-0.18}} &
\cellD{95.17}{\dG{+0.18}} &
\cellD{89.09}{\dR{-5.90}} \\
% \midrule

% \multirow{4}{*}{\textbf{Qwen3-VL-32B}}
% & \textbf{KE} &
% \cellB{64.85} &
% \cellD{65.33}{\dG{+0.48}} &
% \cellD{63.89}{\dR{-0.96}} &
% \cellD{60.15}{\dR{-4.70}} &
% -- \\

% & \textbf{KM} &
% \cellB{65.38} &
% \cellD{66.17}{\dG{+0.79}} &
% \cellD{65.28}{\dR{-0.10}} &
% \cellD{63.56}{\dR{-1.82}} &
% -- \\

% & \textbf{IFE} &
% \cellB{25.52} &
% \cellD{25.93}{\dG{+0.41}} &
% \cellD{25.24}{\dR{-0.28}} &
% \cellD{25.29}{\dR{-0.23}} &
% -- \\

% & \textbf{RE\_Math} &
% \cellB{74.36} &
% \cellD{75.08}{\dG{+0.72}} &
% \cellD{74.10}{\dR{-0.26}} &
% \cellD{74.79}{\dG{+0.43}} &
% -- \\

\bottomrule
\end{tabularx}
}
\caption{\textbf{Hallucination Rates Under Different Shot Settings.}
All main values are in \% (omitted). Deltas for each model are relative to its own baseline (1-shot). Only deltas are colored: \textcolor{red}{red} indicates improvement (lower), and \textcolor{green!60!black}{green} indicates degradation (higher). ``--'' denotes unavailable data.}
\label{tab:shot-ablation-delta-compact}
\end{table}

Overall, common mitigation strategies are better understood as operations that impose bias or reinforcement on specific components. As a result, their benefits are strongly mechanism dependent and may come at the expense of other components. Reasoning SFT is more inclined to repair failures in reasoning integration, but it may amplify biases related to knowledge retrieval or constraint execution. Instruct models tend to improve the stability of instruction execution and alleviate some knowledge conflicts, but this does not equate to stronger multistep reasoning. Meanwhile, simply increasing or decreasing ICL demonstrations provides limited additional knowledge and contextual support, so the marginal effect remains small.

\subsection{Causes of Hallucination Patterns (RQ3)}
% 在这一节中我们通过对LLMs关于输入数据的注意力可视化以及激活神经元数量的统计分析来解释关于Hallucination 的Internal causes
% \paragraph{注意力图}
% 如ref{fig:attention}所示，LLM在KE和KM两种情况下产生幻觉时发生了不同的...

\begin{figure}[ht]
\centering 
\includegraphics[width=1\linewidth]{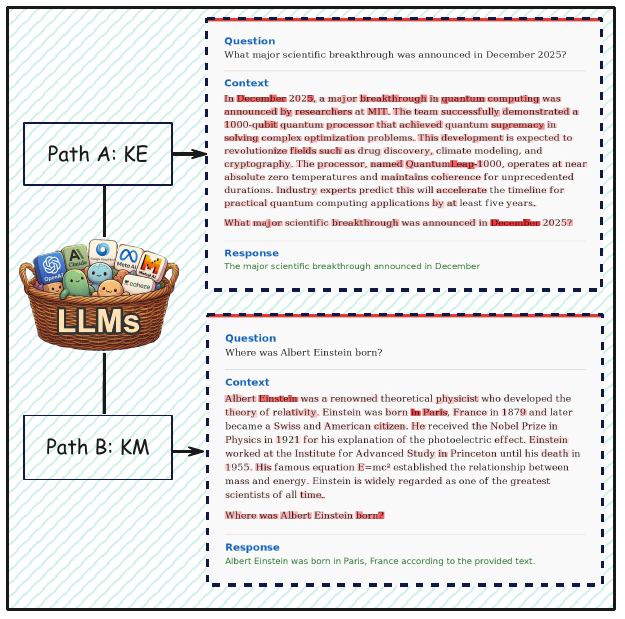} 
\caption{Visualization of Attention Maps in KE and KM} 
\label{fig:attention} 
\end{figure}

To further elucidate the mechanisms underlying model hallucinations in knowledge conflicts, we compare the Attention Maps of KE and KM in Figure~\ref{fig:attention}.

\paragraph{Dominance of Parametric Priors.}
In KE cases, the input provides explicit evidence that contradicts the model’s parametric knowledge. The top of Figure \ref{fig:attention} shows a diffuse attention pattern: the model attends to general entity tokens but does not strongly focus on the specific evidence tokens that determine the correct fact. As a result, strong parametric priors pull the model toward its memorized retrieval and suppress evidence extraction, causing the final answer to drift away from the provided facts.

\paragraph{Dominance of Misleading Context.}
KM cases fail differently. When the model lacks the needed parametric knowledge, or even when the knowledge exists but is not effectively retrieved, misleading context can dominate. The bottom of Figure \ref{fig:attention} shows attention becoming overly concentrated on wrong tokens in the deceptive context. The Qwen3-4B example in Appendix \ref{km}  is consistent with this: instead of detecting an information gap or a conflict, the model treats misleading cues as evidence and produces confident but incorrect conclusions. This implies that without strong internal knowledge, the model is easily misled by false inputs, prioritizing consistency with the context over factual accuracy.

\begin{table}[ht]
\centering

\renewcommand{\arraystretch}{1.3}
\resizebox{0.48\textwidth}{!}{
\begin{tabular}{llccc}
        \toprule
        \textbf{Model} & \textbf{Task Type} & \textbf{ActN} & \textbf{AttnSp} & \textbf{GFLOPs} \\
        \midrule
        \multirow{2}{*}{\textbf{Qwen3-4B}} 
        & IFE (Non-Refusal) & 7.26 M & 0.0191 & 215.7 \\
        & IFE (Refusal)     & 7.72 M & 0.0175 & 229.3 \\
        \midrule
        \multirow{2}{*}{\textbf{LLama3.1-8B}} 
        & IFE (Non-Refusal) & 10.41 M & 0.0148  & 468.9 \\
        & IFE (Refusal)     & 11.06 M & 0.0137 & 498.4 \\
        \midrule
        \multirow{2}{*}{\textbf{Qwen3-1.7B}} 
        & IFE (Non-Refusal) & 4.73 M & 0.0208 & 91.7 \\
        & IFE (Refusal)     & 5.03 M & 0.0191 & 97.5 \\
        \midrule
        \multirow{2}{*}{\textbf{Qwen3-14B}} 
        & IFE (Non-Refusal) & 13.58 M & 0.0169 & 755.0 \\
        & IFE (Refusal)     & 14.44 M & 0.0154 & 802.6 \\
        \bottomrule
    \end{tabular}

}
\caption{IFE Task Comparison on different answer, Abbreviations: ActN = active neurons; AttnSp = attention sparsity; GFLOPs = giga floating-point operations.}
\label{tab:ife_comparison}
\end{table}

\paragraph{Computational Cost of Refusal.}
Table \ref{tab:ife_comparison} quantifies the computational cost of getting this right. Across all four models, Qwen3-4B, Llama3.1-8B, Qwen3-1.7B, and Qwen3-14B, the refusal state in the IFE task consistently uses more ActN and higher GFLOPs than the non-refusal state. Refusal also shows lower attention sparsity, suggesting the model must attend to more signals and perform heavier cross-checking to identify conflicts and suppress hallucinations, whereas generating a hallucinated answer can be computationally cheaper.

\section{Conclusion}
We introduce \textbf{\textit{PRISM}}, a benchmark designed to evaluate hallucination dimensions based on the response pipeline of LLMs. By decomposing the hallucination phenomenon into four dimensions known as KE, KM, RE, and IFE, we achieve identification of the stages where models fail. Extensive experiments demonstrate that while proprietary models generally outperform open-source models, all models exhibit significant trade-offs across different dimensions. Existing mitigation strategies function essentially as inductive biases imposed on specific generation stages. These approaches often improve performance in one dimension while compromising the stability of memory retrieval or logical reasoning. \textbf{\textit{PRISM}} provides a quantitative basis for understanding these complex interactions and lays a solid foundation for future model selection, targeted training optimization, and trustworthy hallucination governance.

\section{Limitations}

Although \textbf{\textit{PRISM}} provides a robust framework for hallucination diagnosis, this study still has several limitations. First, \textbf{\textit{PRISM}} currently focuses solely on diagnosing hallucinations in the text modality and does not yet cover cross-modal hallucination challenges introduced by visual and auditory information in multimodal large language models. Meanwhile, the benchmark concentrates on five high-resource languages, and its applicability to low-resource linguistic settings remains to be validated. Second, real-world knowledge is continuously evolving, and static evaluation sets struggle to fully capture knowledge updating and forgetting when models encounter real-time information. Although our OOD temporal knowledge data spans March--November 2025 and the training cutoffs of most evaluated models predate this window, we cannot entirely rule out the risk of data contamination for models whose training boundaries are undisclosed. Furthermore, the benchmark relies on authoritative static corpora to ensure verifiability and reproducibility, which may result in insufficient coverage of long-tail information. 

Finally, \textbf{\textit{PRISM}} is intended as a research diagnostic tool and should not serve as the sole basis for assessing model safety in high-stakes domains such as healthcare and law. Future work will extend evaluation to multimodal and multilingual settings, introduce rolling temporal window updates and systematic contamination detection, to further enhance the benchmark's practicality and robustness.

\section*{Acknowledgments}
This work was supported by the National Natural Science Foundation of China (72442025). We thank the anonymous participants for taking part in our study. We are also grateful to the members of the DUFE Fintech Lab for their helpful comments.

\bibliography{custom}

\newpage

\appendix
\section*{Appendix}
\section{Benchmark Comparisons}
\label{app:benchmark_comparisons}

Table~\ref{Long_comparison}  provides a comprehensive comparison between \textbf{\textit{PRISM}} and existing benchmarks in \textit{Evaluation Scope} and \textit{Methodological Design}. Most existing benchmarks concentrate on only certain aspects of hallucination, failing to provide a holistic examination of failures. In terms of the depth of the evaluation, they rely largely on posterior analysis, making the attribution of hallucination causes unclear.

In contrast, \textbf{\textit{PRISM}} introduces a diagnostic framework that covers four distinct failure mechanisms. By defining hallucination categories upfront at the input level, \textbf{\textit{PRISM}} enables a more accurate identification of where failures occur and why they arise.

\begin{table*}[h]
\centering
\small

% \setlength{\tabcolsep}{5pt}
% \renewcommand{\arraystretch}{1.25}
% \resizebox{\textwidth}{!}{
\begin{tabular}{l c l c c c c}
\toprule
\multicolumn{1}{c}{\multirow{2}{*}{\textbf{Benchmark}}} &
\multicolumn{4}{c}{\textbf{Evaluation Scope}} &
\multicolumn{2}{c}{\textbf{Methodological Design}} \\

\cmidrule(lr){2-5}
\cmidrule(lr){6-7}

& \textbf{KE} & \textbf{KM} & \textbf{RE} & \textbf{IFE} & \textbf{Variable Control} & \textbf{Diag. Mode} \\
\midrule

TruthfulQA \citep{lin2022truthfulqa} & \cmark &  \xmark
&\xmark & \xmark
& \xmark & \xmark
\\

HaluEval \citep{li2023halueval} & \cmark &  \cmark
&\cmark & \xmark
& \xmark & \xmark
\\

FActScore \citep{min2023factscore} & \cmark &  \cmark
&\xmark & \xmark
& \xmark & \xmark
\\
FactCHD \citep{chen2023factchd} & \cmark &  \xmark
&\cmark & \xmark
& \xmark & \xmark
\\

FELM \citep{chen2023felm} & \cmark &  \xmark
&\cmark & \xmark
& \xmark & \xmark
\\
HalluQA \citep{cheng2023evaluatinghallucinationschineselarge} & \cmark & \xmark & \xmark & \cmark & \xmark & \cmark \\
FreshQA \citep{vu2023freshllms} & \cmark &  \cmark
&\xmark & \xmark
& \cmark & \xmark
\\

FollowBench \citep{jiang2024followbench} & \xmark &  \xmark
&\cmark & \cmark
& \xmark & \xmark
\\

SimpleQA \citep{wei2024measuring} & \cmark & \cmark & \xmark & \xmark & \xmark & \xmark \\
Collu-Bench \citep{jiang2024collu} & \xmark & \xmark & \cmark & \cmark & \cmark & \xmark \\
LongFact \citep{wei2024long} & \cmark & \xmark & \xmark & \xmark & \xmark & \xmark \\
RAGTruth \citep{niu2024ragtruth} & \cmark & \xmark & \cmark & \xmark & \xmark & \xmark \\
DiaHalu \citep{chen2024diahalu}& \xmark& \xmark& \cmark& \cmark& \xmark&\xmark\\
LogicAsker \citep{wan2024logicasker} & \xmark & \xmark & \cmark & \xmark & \cmark & \cmark \\
RefuteBench \citep{yan2024refutebench} & \xmark & \xmark & \cmark & \cmark & \xmark & \xmark \\
HaluEval-Wild \citep{zhu2024halueval} & \cmark & \cmark & \cmark & \xmark & \xmark & \cmark \\
REFCHECKER \citep{hu2024refchecker} & \xmark & \xmark & \cmark & \cmark & \xmark & \xmark \\
ERBench \citep{oh2024erbench} & \xmark & \xmark & \cmark & \cmark & \xmark & \xmark \\
ToolBeHonest \citep{zhang2024toolbehonest} & \xmark & \xmark & \cmark & \cmark & \xmark & \xmark \\
Defan \citep{rahman2024defan} & \cmark & \xmark & \cmark & \cmark & \xmark & \xmark \\
UHGEval \citep{liang2024uhgeval} & \cmark & \xmark & \cmark & \xmark & \xmark & \xmark \\
WildHallucinations \citep{liang2024uhgeval} & \xmark & \cmark & \cmark & \xmark & \xmark & \xmark \\
TofuEval \citep{tang2024tofueval} & \cmark & \cmark & \cmark & \xmark & \xmark & \xmark \\
FavaBench \citep{mishra2024finegrainedhallucinationdetectionediting} & \cmark & \cmark & \cmark & \xmark & \xmark & \xmark \\
ANAH \citep{ji2024anahanalyticalannotationhallucinations} & \cmark & \cmark & \cmark & \xmark & \xmark & \xmark \\
Chinese SimpleQA \citep{he2025chinese} & \cmark & \cmark & \xmark & \xmark & \xmark & \xmark \\
HalluMix \citep{emery2025hallumix} & \xmark & \cmark & \cmark & \xmark & \xmark & \xmark \\
MedHallu \citep{pandit2025medhallu} & \cmark & \cmark & \xmark & \xmark & \xmark & \xmark \\
HalluVerse25 \citep{abdaljalil2025halluverse25finegrainedmultilingualbenchmark} & \cmark & \xmark & \cmark & \cmark & \xmark & \xmark \\
ReasonIF \citep{kwon2025reasonif} & \xmark & \xmark & \cmark & \cmark & \xmark & \cmark \\
MultiHal \citep{lavrinovics2025multihalmultilingualdatasetknowledgegraph} & \cmark & \xmark & \cmark & \xmark & \xmark & \cmark \\
DynaQuest \citep{lin2025dynaquest} & \cmark & \cmark & \xmark & \xmark & \cmark & \xmark \\
CodeHalu \citep{tian2025codehalu}& \cmark& \xmark& \cmark& \xmark& \xmark &\cmark\\
FaithBench \citep{bao2025faithbench}& \xmark& \xmark& \cmark& \cmark& \xmark &\xmark \\
HalluciNot \citep{paudel2025hallucinothallucinationdetectioncontext}& \cmark& \xmark& \cmark& \xmark& \xmark &\xmark \\
HALoGEN \citep{ravichander2025halogen} & \cmark &  \cmark
&\cmark& \xmark
& \xmark & \xmark\\

HalluLens \citep{bang2025hallulens} & \cmark &  \cmark
&\cmark & \xmark
& \xmark & \xmark
\\

\midrule
\rowcolor{aclblue!10}
\textbf{\textit{PRISM (Ours)}} & \cmark &  \cmark
&\cmark & \cmark
& \cmark~& \cmark
\\

\bottomrule
\end{tabular}

% }

% 
\caption{Systematic Comparison of State-of-the-Art Hallucination and Trustworthiness Benchmarks (2022--2025). }
\label{Long_comparison}
\caption*{\footnotesize \textbf{\textit{Note:}} \textbf{Variable Control} denotes whether a benchmark enforces orthogonal isolation of causal factors, ensuring attribution correctness by preventing confounding effects when decomposing hallucination sources.}
% \label{tab:Long_comparison}
\end{table*}

\section{Related Works}
\subsection{Evolution of Hallucination Categorization}

Early research typically classified hallucinations into general categories, such as Intrinsic versus Extrinsic types based on their relationship to the input \citep{ji2023survey}, or distinguished between Factuality and Faithfulness errors based on consistency with external knowledge \citep{zhang2023siren, huang2023survey}. With the expansion of model capabilities in 2024, recent studies have extended these definitions to specific domains. For instance, \cite{ho2024legal} analyzed premise-compliance errors in legal tasks, where models erroneously validate incorrect user assumptions. Furthermore, theoretical studies \citep{sahoo2024comprehensive, suzuki2025hallucinations} suggest that hallucinations are inherent statistical features of probabilistic models rather than simple engineering defects.

However, existing studies mostly focus on the manifestation of errors rather than the root causes. Current frameworks often fail to distinguish whether a failure stems from missing data, flawed reasoning, or an inability to follow instructions, thereby hindering the accurate attribution of errors in complex generation pipelines.

\subsection{Benchmarks for Hallucination Evaluation}

In response to the challenge of hallucination in LLMs, previous studies have developed a series of benchmarks to evaluate hallucinations. Early efforts primarily assessed the model's grasp of static world knowledge and its resilience to interference. Specifically targeting hallucinations, HaluEval \citep{li2023halueval} generates and screens a large number of hallucination samples to evaluate whether LLMs can identify hallucination issues. As understanding of hallucination mechanisms deepened, the evaluation paradigms shifted toward higher granularity. FActScore \citep{min2023factscore} introduced an atomic-level evaluation, dividing long-form text into individual facts to verify their support against reliable knowledge sources. Similarly, FELM \citep{chen2023felm} adopted segment-level annotation, significantly improving the precision of error localization across diverse domains. Recent frameworks have further systematized the definition of errors: HalluLens \citep{bang2025hallulens} formalized the distinction between extrinsic hallucinations (contradicting training data or reality) and intrinsic hallucinations (deviating from input context), while FollowBench \citep{jiang2024followbench} and HALoGEN \citep{ravichander2025halogen} expanded the evaluation scope to include instruction following failures and source-based error attribution, distinguishing memory distortion from fabrication.

Despite these advances, as shown in Table \ref{Long_comparison}, existing benchmarks suffer from a fundamental methodological limitation: their reliance on outcome-oriented and posterior evaluation. \textbf{\textit{PRISM}} is designed to address this critical gap. Unlike prior works, \textbf{\textit{PRISM}} establishes a diagnostic framework based on active probing, isolating the distinct cognitive stages of source memory retrieval, reasoning, and instruction following to precisely locate the origin of hallucinations within the generative pipeline.

\subsection{Strategies for Hallucination Mitigation}
Numerous studies have proposed diverse approaches to mitigate hallucinations in LLMs \citep{tonmoy2024comprehensive}. From the model generation perspective, existing hallucination mitigation methods target hallucination mechanisms at different stages, primarily memory, reasoning, and instruction-following \citep{li2025mitigating}. Memory-stage hallucinations are commonly addressed through knowledge-centric approaches, including retrieval-augmented generation \citep{ayala2024reducing} and knowledge base querying \citep{pan2024unifying}, which aim to alleviate errors caused by missing or incorrect knowledge. Reasoning-stage hallucinations, stemming from inconsistencies in multi-step reasoning, are mitigated by structuring the reasoning process via techniques such as Chain-of-Thought prompting \citep{zhang2024chain} and self-consistency reasoning \citep{liu2025enhancing}, often combined with causal learning \citep{kiciman2023causal} or contrastive learning \citep{xu2023temporal} to improve reasoning robustness. Instruction-stage hallucinations, characterized by deviations from user instructions or task constraints, are typically mitigated through constrained prompting \citep{wu2025mitigating} and post-hoc verification mechanisms, including self-reflection \citep{ji2023towards} and self-consistency checking \citep{liang2024internal}, to enhance instruction-following behavior.

While prior work studies hallucinations from multiple sources, systematic evaluation across different hallucination problems is limited \citep{dang2025survey,chakraborty2025hallucination}. Therefore we introduce \textbf{\textit{PRISM} }and use it to analyze mitigation strategies across hallucinations caused by memory, reasoning, and instruction-following failures.

\section{Data Source and Task Definitions}
\label{source}

In this section, we present the comprehensive taxonomy of the \textbf{\textit{PRISM}} benchmark, detailing the specific failure mechanisms associated with each cognitive stage. To facilitate a granular diagnosis of model hallucinations, we categorize failures into four primary dimensions: KE, KM, RE, and IFE. Each dimension is further divided into specific subcategories to capture distinct error patterns ranging from factual distortions to procedural breakdowns. Table \ref{tab:re_taxonomy} provides rigorous definitions for these subcategories and maps them to their respective data sources, illustrating how our multi-source construction strategy ensures broad coverage across the entire spectrum of potential failures.

For the temporal OOD portion of KM, our TimeOOD data were collected between March and November 2025, which is strictly later than the officially disclosed knowledge cutoffs or public time-boundary information of several representative models evaluated, including Llama~4 Scout \cite{llama_2025}, the Grok~4 series \cite{grok_2025}, and Gemini~3 Pro \cite{gemini_2026}.

\begin{table*}[htbp]
\centering
\small
\begin{tabular}{
>{\raggedright\arraybackslash}m{0.24\linewidth}
>{\raggedright\arraybackslash}m{0.44\linewidth}
c c c c c
}
\toprule
\multicolumn{1}{>{\centering\arraybackslash}m{0.24\linewidth}}{\textbf{Subcategory}} &
\multicolumn{1}{>{\centering\arraybackslash}m{0.44\linewidth}}{\textbf{Definition}} &
\multicolumn{5}{c}{\textbf{Source Category}} \\
\cmidrule(lr){3-7}
 &  & \textbf{EB} & \textbf{ED} & \textbf{RWK} & \textbf{HE} & \textbf{SD} \\
\midrule
\rowcolor{aclblue!10}
\multicolumn{7}{c}{\textbf{Knowledge Error(KE)}}\\
\midrule

\textbf{KE1: Factual Distortion (FD)} &
The model generates factually incorrect outputs despite possessing relevant knowledge, due to errors in knowledge representation, updating or retrieval.
& \cmark &  & \cmark &  \\

\textbf{KE2: Intra-Memory Conflict (IMC) } &
The model internally stores multiple conflicting versions of the same fact, resulting in inconsistent or contradictory answers across different queries, contexts, or interaction turns.
& \cmark & \cmark &  &  &  \\

\textbf{KE3: Entity-Identity Confusion (EIC)} &
The model fails to correctly distinguish between entities with identical names, multiple meanings, or semantic proximity, incorrectly transferring or binding knowledge from one entity to another.
& \cmark & \cmark & \cmark &  &  \\
\midrule

\rowcolor{aclblue!10}
 \multicolumn{7}{c}{\textbf{Knowledge Missing(KM)}}\\
\midrule
\textbf{KM1: Domain-Specific Knowledge (DSK)} &
The model lacks newer or niche knowledge in certain specific domains, which prevents it from answering related questions.
& \cmark &\cmark  &\cmark  &  &  \\

\textbf{KM2: Fictional Knowledge (FK)} &
The model's dataset does not include any knowledge of the fictional content in the question, which prevents the model from answering the relevant question.
&  &  &\cmark  &  &\cmark  \\

\textbf{KM3: Timely Knowledge (TK)} &
The model's dataset does not include content that changes continuously over time or recent events, which prevents the model from answering related questions.
&  &  &\cmark  &  &\cmark  \\

\textbf{KM4: Non-Public Knowledge (NPK)} &
When the model asks questions involving personal thoughts or private information, the dataset may not contain the question content or may refuse to answer due to privacy concerns, thus preventing the model from answering the relevant questions.
&  &  &\cmark  &  &\cmark  \\

\midrule
\rowcolor{aclblue!10}
 \multicolumn{7}{c}{\textbf{Reasoning Error(RE)}}\\
\midrule

\textbf{RE1: Logical Fallacy (LF)} &
The model fails to adhere to abstract formal logic or causal rules, deriving invalid conclusions from valid premises.
& \cmark & \cmark &  & \cmark &  \\

\textbf{RE2: Procedural Failure (PF)} &
The model fails to maintain the correct sequence or continuity in multi-step or multi-hop tasks, resulting in lost chains of thought or skipped intermediate steps.
& \cmark & \cmark &  &  &  \\

\textbf{RE3: Information Integration Failure (IIF)} &
The model fails to identify, prioritize, or synthesize scattered information from long or noisy contexts, leading to reasoning based on incomplete evidence.
& \cmark &  & \cmark &  & \cmark \\

\textbf{RE4: Mathematical Reasoning Failure (MRF)} &
Errors in mathematical or algorithmic reasoning involving derivation, calculation, or implementation.
& \cmark &  &  & \cmark &  \\
\midrule
\rowcolor{aclblue!10}
 \multicolumn{7}{c}{\textbf{Instruction Following Error(IFE)}}\\
\midrule

\textbf{IFE1: Explicit Format (EF)} &
The model fails to adhere to strict structural specifications or text syntactic patterns, resulting in invalid data formats or template violations.
& \cmark & & & & \\

\textbf{IFE2: Length Constraints (LC)} &
The model violates quantitative length requirements, failing to keep the output within the specified word count or boundary limits.
& \cmark & \cmark & & & \\

\textbf{IFE3: Language Constraints (LgC)} &
The model fails to respond in the specified target language, or incorrectly mixes multiple languages when a single language is required.
& \cmark & \cmark & & & \\

\textbf{IFE4: Complex \& Cognitive Load (CCL)} &
The model fails to parse or execute complex instructions involving negation or conditional logic, often ignoring critical constraints or specific rule details.
& \cmark & & & \cmark & \\

\bottomrule
\end{tabular}
\caption{Taxonomy of our benchmark and corresponding data sources. 
\textbf{EB}: Existing Benchmarks; 
\textbf{ED}: Enhanced Datasets derived from existing benchmarks; 
\textbf{RWK}: Real-world Knowledge collected from authoritative sources; 
\textbf{HE}: Human Exams serving as human-level reasoning references; 
\textbf{SD}: Synthetic Data generated under controlled constraints.}
\label{tab:re_taxonomy}
\end{table*}

\section{Construction Pipeline}
\label{pipline}

To ensure high data quality, we implement a rigorous four-stage construction pipeline. As summarized in Table \ref{tab:appendix_pipeline_stats}, this process filters an initial pool of over 33,000 candidate samples down to the final 9,448 high-quality instances in the PRISM benchmark. Full implementation details for each stage are provided in the following subsections.

\begin{table*}[htbp]
\centering
\resizebox{\linewidth}{!}{
\begin{tabular}{l l r r r r r}
\toprule
\small
\textbf{Stage} & \textbf{Action} & \textbf{KE} & \textbf{KM} & \textbf{RE} & \textbf{IFE} & \textbf{Total} \\
\midrule
Data Collection & Mining \& Synthesis & 4,982 & 8,117 & 10,696 & 9,539 & 33,334 \\
Data Cleaning & Conversion \& Denoising & 4,200 & 6,824 & 9,075 & 7,943 & 28,042 \\
Multi-agent Construction & Evidence Grounding \& Quality Scoring & 2,655 & 2,979 & 4,638 & 4,310 & 14,582 \\
Human Selection & Expert Adjudication & 1,933 & 2,078 & 2,995 & 2,442 & 9,448 \\
\midrule
\textbf{Pass Rate} & \textbf{Final / Initial} & \textbf{38.80\%} & \textbf{25.60\%} & \textbf{28.00\%} & \textbf{25.60\%} & \textbf{28.34\%} \\
\bottomrule
\end{tabular}
}
\caption{Per-stage sample counts and pass rates of the PRISM construction pipeline}
\label{tab:appendix_pipeline_stats}
\end{table*}

\subsection{Data Preprocessing}
\paragraph{Data Conversion.}
Our self-constructed raw materials include both web-based text and PDF documents. To unify downstream processing, we convert materials from different sources into a sequential Markdown representation. For web-based text, we use Trafilatura \citep{barbaresi2021trafilatura} for main-content extraction; this tool is designed for boilerplate removal in web documents and effectively separates core content from page noise. For PDF files, we use MinerU \citep{wang2024mineru,niu2025mineru2}. This tool understands complex page layouts, allowing it to accurately reconstruct the document's structure and maintain the correct reading flow.

\paragraph{Data Cleaning.}
After document conversion, we apply rule-based data cleaning to remove noise that is not directly relevant to evaluation instance construction. First, we eliminate overlapping or highly similar text segments to reduce redundancy and prevent repeated information from affecting downstream processing. Second, we remove in-text citation markers, footnotes, and external links, and discard descriptive or metadata content such as copyright notices, navigation cues, and residual formatting artifacts. Finally, text segments that are clearly incomplete, lack sufficient context, or cannot be understood independently are removed at this stage. Based on these steps, we obtain a consistent and readable raw corpus.

\paragraph{Corpus Formation.}
We construct pre-defined text units as the basic processing granularity for downstream multi-agent workflows. Prior work shows that full-document processing leads to overly long contexts and poor localization, while paragraph-level inputs often lack sufficient context; semantically coherent text units offer a better trade-off between contextual completeness and efficiency \citep{liu2021dense, liu2024lost}. Accordingly, after document conversion and data cleaning, text units are built from section hierarchy and paragraph boundaries, merging adjacent paragraphs when necessary and applying token-level length constraints. For web documents without explicit structure, units are formed using paragraph boundaries. Each unit preserves source identifiers and positional metadata to enable precise localization and traceability.

\subsection{Multi-agent Data Construction}
Generating complex evaluation samples in a single pass is often unreliable because models struggle to follow multiple rules at the same time  \citep{wei2022chain}. To address this, we propose a four-agent framework that breaks the construction process into smaller steps. Unlike black box generation, our method transforms raw text into structured data stage by stage \citep{khot2023decomposed}. We assign specific roles to separate agents \citep{wu2023autogen}. This ensures that each instance includes a clear question, valid evidence from the source, and a precise error label. This modular design automates the work and makes it easier to fix or improve specific parts of the process.

\paragraph{Schema Normalizer Agent.}
The Schema Normalizer Agent converts the raw corpus into a unified standard format. It separates the question, answer, and source information, without adding new facts. The standardized output then serves as a basis for the Evidence Retriever Agent, allowing it to focus on finding relevant supporting materials. The prompt used for this agent is shown in Appendix~\ref{app:Prompts for Tasks} \hyperref[box:SchemaNormalizer_prompt]{Prompt for Schema Normalizer}.

\paragraph{Evidence Retriever Agent.} The Evidence Retriever Agent ensures that every data instance is supported by clear evidence. We use Gemini 3 Pro, known for its strong long-context capabilities, to power this agent. It searches the raw corpus for specific text segments that validate the connection between the question and the answer. Any instance that relies on hidden assumptions or cannot be traced to a clear source in the text is removed at this stage. This ensures that the retained data is fully grounded, preventing the evaluation from relying on unverifiable background knowledge \citep{menick2022teaching,bohnet2022attributed}. The details of its prompt are in Appendix~\ref{app:Prompts for Tasks} \hyperref[box:Retriever_prompt]{Prompt for Evidence Retriever}.

\paragraph{Type Classifier Agent.}
The Type Classifier assigns each verified example to one of four failure dimensions. It reads the question and the reference answer, identifies what the example mainly tests, such as source memory, reasoning, or following instructions, and assigns one label. This step does not change the content. It only groups examples by dimension so we can report where models tend to fail. Appendix~\ref{app:Prompts for Tasks} \hyperref[box:TypeClassifier_prompt]{Prompt for Type Classifier} presents the detailed prompt design.

\paragraph{Quality Scoring Agent.}
The Quality Scoring Agent evaluates each instance across multiple dimensions, focusing on factuality, discriminability, and clarity. Based on these criteria, the agent assigns a quality score to each instance. The primary goal of this step is to filter out low-quality data: only high-scoring instances are retained, while those that are trivial are discarded. This rigorous screening ensures that only challenging instances proceed to the human selection stage. The detailed prompt structure is presented in Appendix~\ref{app:Prompts for Tasks} \hyperref[box:QualityScoring_prompt]{Prompt for Quality Scoring}.

\subsection{Human Selection}

We propose an expert-driven, three-dimensional quality selection strategy for \textbf{\textit{PRISM}} candidate samples. This strategy targets two task families formed by four dimensions in the benchmark, namely knowledge-related and reasoning/instruction-related tasks. The core goal is to ensure that the final item bank satisfies Clarity, Domain Relevance, and Coherence. The selection process proceeds in two stages: (i) stratified purposive sampling and expert recruitment; (ii) multi-round review and filtering based on the three dimensions.

\paragraph{Expert Recruitment.}
We use stratified purposive sampling to form two non-overlapping expert review teams, and each team is confined to its corresponding task family to reduce the risk of cross-mechanism misjudgment and evaluation standard drift \citep{palinkas2015purposeful,stratton2024purposeful}. We recruit a total of $N=80$ reviewers, including:
\begin{itemize}
  \item \textbf{Panel A: Knowledge \& Source Panel. $n=40$.} Responsible for KE and KM samples. The core responsibility is to judge, without introducing external background knowledge, whether an item truly relies on parametric knowledge.
  \item \textbf{Panel B: Reasoning \& Instruction Panel. $n=40$.} Responsible for RE and IFE samples. The core responsibility is to ensure that any prerequisite knowledge involved in the sample is assumed to be known to the model, that the reasoning chain can be closed within the item itself, and that instruction constraints are clear and executable.
\end{itemize}

Candidates are sourced from universities and research institutes, NLP/IR/reasoning evaluation research communities, industry research and evaluation teams, and professional technical communities. Eligibility and screening criteria are set by panel: the Knowledge \& Source Panel requires an advanced degree or three years or more of work experience in fact checking or information retrieval; the Reasoning \& Instruction Panel requires three years or more of work experience in reasoning tasks or instruction specification development.

\paragraph{Filtering Rules.}
To implement our dual-track validation framework, we conduct a multi-round expert review process. Each candidate item is evaluated against a predefined set of track-specific quality dimensions. Experts provide independent scores and brief rationales, and items are assigned one of three decisions: retain, remove, or revise then re-review. This iterative process continues until convergence, ensuring that every item in the final \textbf{\textit{PRISM}} bank meets our quality baseline. The following sections detail the evaluation dimensions.

\begin{itemize}
  \item \textbf{Clarity:}
  Clarity is the foundation of reliable selection. Once an item is written vaguely, a model's score can be shifted by reading habits and by how it fills in missing premises, rather than by the ability differences we care about. Bandalos \cite{Bandalos2018} points out that ambiguity introduced by item wording can introduce construct-irrelevant variance, thereby harming reliability and validity. In benchmark construction, clarity is also directly tied to face validity, namely whether readers can immediately see what the item is asking, whether the information is written explicitly, and whether answering does not require guessing the author's intent. Only when clarity is met do the subsequent judgments of relevance and coherence become meaningful.\\
  \textbf{Core Question:} \textit{Is the item statement precise, or could readers arrive at two equally reasonable but different interpretations?}

  \item \textbf{Domain Relevance:}
  Domain relevance means that an item genuinely operates on the dimensions we define, rather than being driven by side issues. Haynes et al.\ \citep{Haynes1995} summarize the key points of content validity in two aspects: the content should stay close to what is intended to be measured, and the coverage should be representative. In human selection, this means avoiding items that appear to test a certain dimension but end up mainly testing something else, for example, succeeding through obscure background knowledge, or creating differences through wording tricks. Items with strong domain relevance are less likely to be derailed by test-gaming patterns, and they better support interpreting errors as deficiencies in a specific dimension, which is consistent with the testing standards that emphasize evidence for score interpretation and use \citep{Messick1995}.\\
  \textbf{Core Question:} \textit{Does the main difficulty of this item primarily come from the dimension it is labeled with, rather than from irrelevant factors?}

  \item \textbf{Coherence:}
  Coherence requires internal consistency and a closed information chain. The information provided in the prompt should support the reference answer or label; the item should not present one story in the material and another in the answer, and it should not contain facts that conflict across sentences. Kane's \citep{Kane2013} argument-based approach to validation emphasizes that any interpretation relies on a chain of inferences; if the internal chain of an item breaks, it effectively pushes key inferences onto the model or the evaluator, making conclusions unstable. Similarly, FEVER binds support or refute judgments to necessary evidence so that the judgment does not drift away from the evidence \citep{Thorne2018}. We include coherence as a dimension to ensure that each sample can self-consistently explain why the answer is what it is and why the label is what it is.\\
  \textbf{Core Question:} \textit{Using only the information provided in the item, can one stably reach the same conclusion, and is the answer or label consistent with the material?}
\end{itemize}

\section{Question Quality Filter}
\subsection{Sample Quality Evaluation Dimension}
We argue that a high-quality QA instance should exhibit factual correctness, unambiguous categorical targeting, and linguistic clarity. Accordingly, we define three scoring dimensions to evaluate sample quality: A (Factuality), B (Discriminability), and C (Clarity). More detailed criteria for each scoring dimension are provided in Table~\ref{tab:quality_criteria}. Below, we outline the purpose and rationale of each dimension.

\begin{table*}[h]
\centering
\small
\begin{tabular}{p{0.08\linewidth} p{0.86\linewidth}}
\toprule

\rowcolor{aclblue!10}
\multicolumn{2}{l}{\textbf{A. Factuality}} \\
\midrule
1-2 & Largely incorrect or contradictory to the source with no supporting source. \\
3-4 & Core claims unsupported or inconsistent with the source with at most one weakly related source. \\
5-6 & Some alignment with the source with main claims partially supported by one source. \\
7-8 & Minor inaccuracies with main claims supported by two sources.\\
9-10 & Fully faithful to the source with all claims supported by more than two sources. \\

\midrule
\rowcolor{aclblue!10}
\multicolumn{2}{l}{\textbf{B. Discriminability}} \\
\midrule
1-2 & Absolutely unclassifiable. \\
3-4 & Features of multiple hallucination categories present. \\
5-6 & Dominant hallucination mechanism not immediately evident.\\
7-8 & Largely attributable to a primary hallucination category. \\
9-10 & No cross-category features. \\

\midrule
\rowcolor{aclblue!10}
\multicolumn{2}{l}{\textbf{C. Clarity}} \\
\midrule
1-2 & Disorganized, vague, and difficult to interpret. \\
3-4 & Noticeable ambiguity or redundancy. \\
5-6 & Generally understandable.\\
7-8 & Clear and fluent, with minor room for refinement.\\
9-10 & Highly clear and precise, immediately interpretable. \\

\bottomrule
\end{tabular}
\caption{Evaluation criteria used in our study. Each dimension is rated on a 1--10 scale with detailed scoring guidelines.}
\label{tab:quality_criteria}
\end{table*}

\paragraph{Factuality.}
Factuality evaluates whether an answer is correct and strictly grounded in the original source evidence. Classic benchmarks such as HotpotQA explicitly incorporate factuality by requiring models to answer multi-hop questions while identifying supporting evidence \citep{yang2018hotpotqa}. Subsequent results on TruthfulQA show that even large-scale language models frequently produce imitative falsehoods, highlighting the importance of evaluating factual correctness \citep{lin2022truthfulqa}. The scoring agent compares each claim in the answer against the original corpus to ensure evidence consistency, preventing the introduction of hallucinated content that is inconsistent with or unverifiable from the source material. A high score indicates that the answer is factual, verifiable, and free from factual errors, logical contradictions, or unsupported assertions.
\paragraph{Discriminability.}
Discriminability measures whether the question precisely targets a single failure category, minimizing label overlap or semantic ambiguity. Baldock et al. \cite{baldock2021deep} introduced the SoftGap metric in OOD detection, showing that a larger margin implies more confident, unambiguous predictions. Plaut et al. \cite{plaut2024probabilities} further validate margin as a reliable uncertainty indicator across QA benchmarks, reinforcing its utility for measuring categorical precision. This margin-based approach ensures the rigor of our discriminability mechanism. Let $s_i$ be the predicted confidence score for category $i$, computed via softmax:
\[
s_i = \frac{\exp(z_i)}{\sum_k \exp(z_k)},
\]
and let $s_{(1)}, s_{(2)}$ denote the highest and second-highest scores respectively. The margin is computed as:
\[
\text{Margin} = \max_i s_i - \max_{j \ne i} s_j.
\]
A larger margin reflects clear category targeting, while a smaller one indicates ambiguity. High discriminative precision ensures each QA pair isolates a specific failure mode, strengthening the benchmark’s effectiveness in identifying the root causes of hallucination.

\paragraph{Clarity.}
Clarity is the basis of valid assessment. We evaluate clarity for the question–answer pair as a whole. This means explicitly naming all entities and context and using precise, direct phrasing in well-structured sentences. When both the question and the answer are clearly phrased, there is effectively a single intended reading, so any factual inconsistency or hallucinated content can be unambiguously identified and traced. Indeed, discourse studies of evasive or ambiguous answers find that unclear responses often contain contradictions, topic shifts, or incomplete fragments, resulting in multiple interpretations \citep{confIJCAI20,thomas2024never}. This approach matches standard QA dataset practices. By enforcing clarity, we avoid these confounds and ensure that hallucination judgments reflect genuine content errors rather than mere linguistic confusion \citep{ouyang2025hoh}.

\subsection{Sample Quality Score}
We adopt a two-step filtering process to automatically assess and refine QA instances before final use. The first step is weighted evaluation, which aggregates factuality, discriminative precision and clarity into a weighted quality score. The second step applies threshold-based elimination to eliminate samples with significant flaws in any individual aspect. Both procedures are implemented by the Quality Scoring Agent to ensure scoring consistency, as detailed in the following paragraphs.
\begin{figure*}[htb]
    \centering
    \includegraphics[width=1\linewidth]{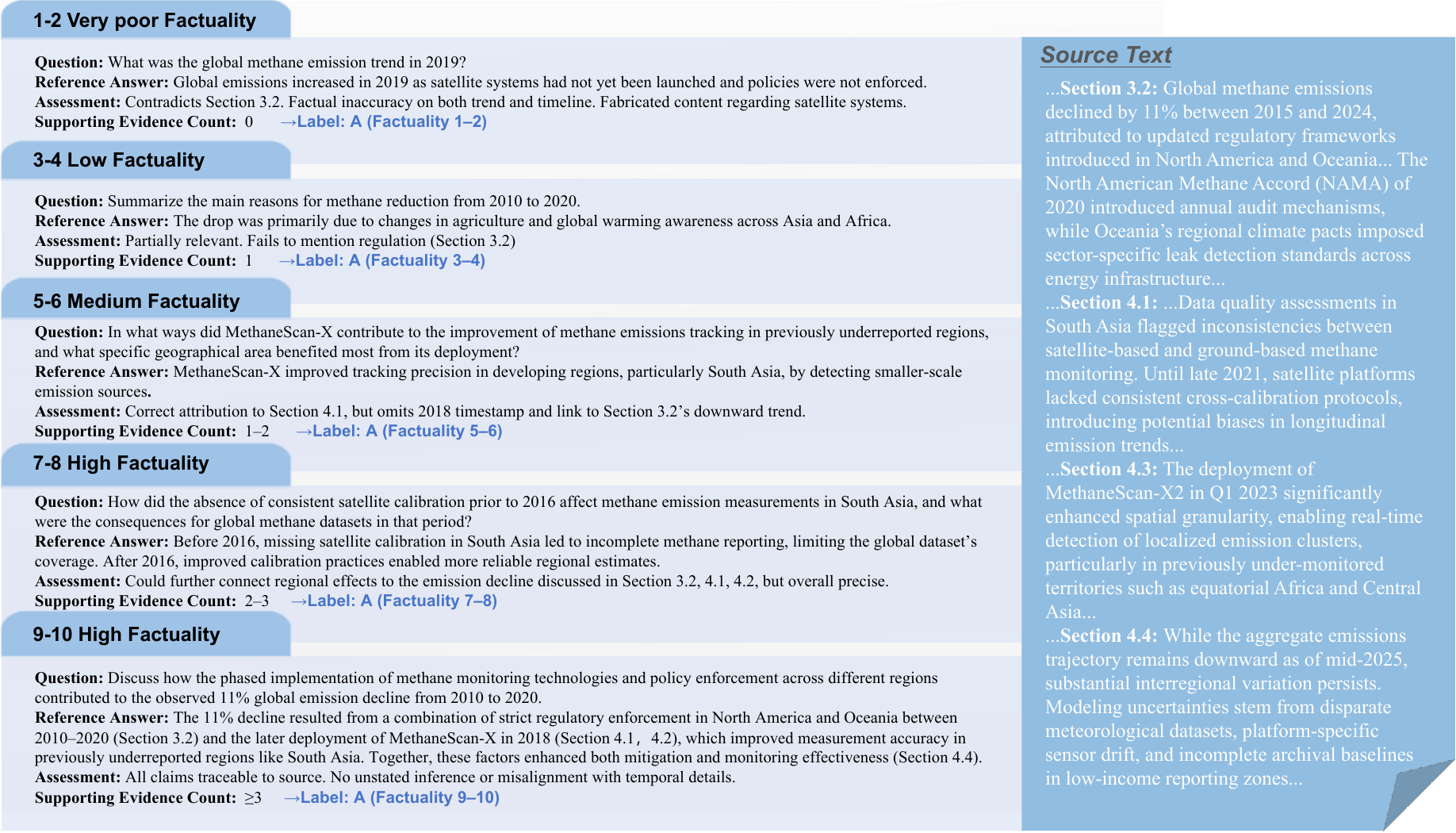}
    \caption{Illustrative Examples of QA Instances Across Factuality Score Bands}
    \label{fig:qualityscore}
\end{figure*}

\paragraph{Scoring Rules and Bands.}
Each question-answer pair is evaluated by a quality agent along three dimensions: factuality, discriminability, and clarity. A score from 1 to 10 is assigned for each dimension, based on explicit scoring bands detailed in Table \ref{tab:quality_criteria}. Agents select the highest level that is fully satisfied and provide a concise justification. To aid transparency and interpretation, Figure \ref{fig:qualityscore} presents QA examples at different factuality levels, using a shared source to illustrate how alignment and evidence grounding influence scoring outcomes.

\paragraph{Weighted Evaluation.}
To calibrate scientifically grounded weights across factuality, discriminability, and clarity, we asked domain experts to compare QA triplets based on benchmark goals. Their pairwise preferences were used to learn scoring weights via weighted aggregation and pairwise loss, guided by the following three dimensions:

\begin{itemize}\setlength\itemsep{0.2em}
  \item Preserve factual consistency with source content.
  \item Enhance discriminability for hallucination mechanism analysis.
  \item Ensure clarity to reduce ambiguity and redundancy.
\end{itemize}

This yields a set of constraints $\mathcal{P} = \{(i, j)\}$, where $(i, j)$ indicates that example $i$ is preferred over $j$.

Given the weighted scoring function:
\[
S(x) = \alpha A(x) + \beta B(x) + \gamma C(x),
\]
we require $S(x_i) > S(x_j)$ for each $(i, j) \in \mathcal{P}$. To enforce this, we adopt a pairwise ranking loss from the learning-to-rank literature:
{\small
\[
\mathcal{L}(\alpha, \beta, \gamma) = \sum_{(i,j) \in \mathcal{P}} \log\left(1 + \exp\left( - \left(S(x_i) - S(x_j)\right) \right)\right).
\]
}

Minimizing this loss corresponds to maximizing the likelihood under a Bradley–Terry preference model \cite{burges2005learning}, where the probability of preferring sample $i$ over $j$ increases with $S(x_i) - S(x_j)$. In practice, we normalize $(\alpha, \beta, \gamma)$ to sum to one and optimize them using gradient descent on expert-labeled triplet preferences. Specifically, 20 experts labeled 500 sample triplets from early benchmark drafts, producing constraints that led to final weights of $(0.5, 0.3, 0.2)$. The higher weight on discriminability highlights its pivotal role in isolating interpretable hallucination types, while factuality contributes to source grounding and clarity supports linguistic precision. Together, these dimensions strengthen the benchmark's ability to support mechanism-level diagnosis and targeted mitigation of hallucination.

\paragraph{Threshold-Based Elimination.}
Although the weighted score provides a global quality estimate, it may overlook critical weaknesses in individual dimensions. We therefore enforce per-dimension thresholds to exclude samples with inadequate factuality, discriminability, or clarity. Let $\mathcal{R}$ be the set of samples rejected by quality agents, and let $\hat{\mathcal{R}}(T)$ denote the set automatically filtered under threshold $T$.
We define rejection recall as:

\[
\text{Recall}_{\text{reject}}(T)
= \frac{|\mathcal{R} \cap \hat{\mathcal{R}}(T)|}{|\mathcal{R}|}.
\]

This recall reflects the consistency between quality scoring agents and threshold-based filtering. Empirical validation shows setting $T = 7.0$ yields a recall of 90\%.  To maintain clarity in category-level analysis, we also enforce that all dimension scores exceed 7.0, removing low-quality QA pairs that could obscure mechanism-level insights.

\section{Evaluation Results}
\label{result_table}

This section provides the complete experimental results regarding the four core evaluation dimensions discussed in RQ1. Tables~\ref{tab:prism-ke} to ~\ref{tab:prism-ife} display the specific results for KE, KM, RE, and IFE, respectively.

% ========================= % KE % =========================

\begin{table*}[h]
\centering
\setlength{\tabcolsep}{4pt}
\renewcommand{\arraystretch}{0.9}
\resizebox{\textwidth}{!}{%
\begin{tabular}{l cccccccccc cc ccc}
\toprule
\multirow{2}{*}{Model}  & \multicolumn{10}{c}{FD} & \multicolumn{2}{c}{IMC} & \multicolumn{3}{c}{EIC} \\
\cmidrule(lr){2-11} \cmidrule(lr){12-13} \cmidrule(lr){14-16}
& Art & Biz & DefAn & Food & Lang & LawCrimMil & Rel & Sci & Sports & Truth & Numeric & Text & SelfBuilt & WhoEnt & WiC \\
\midrule
\multicolumn{16}{l}{\textbf{\textit{Proprietary LLMs}}} \\
GPT-5.2 & 100.00\% & 96.97\% & 16.57\% & 100.00\% & 100.00\% & 97.50\% & 100.00\% & 100.00\% & 97.92\% & 68.15\% & 54.22\% & 92.63\% & 73.98\% & 54.77\% & 79.87\% \\
GPT-5.1 & 97.78\% & 93.94\% & 38.86\% & 100.00\% & 77.78\% & 95.00\% & 97.96\% & 100.00\% & 95.83\% & 65.61\% & 37.35\% & 56.84\% & 45.92\% & 38.59\% & 71.14\% \\
GPT-4o-20241120 & 95.56\% & 90.91\% & 19.43\% & 100.00\% & 81.48\% & 92.50\% & 100.00\% & 96.77\% & 100.00\% & 65.39\% & 55.42\% & 94.74\% & 73.47\% & 73.44\% & 70.47\% \\
Gemini-3-Pro & 100.00\% & 96.97\% & 42.86\% & 100.00\% & 96.30\% & 97.50\% & 100.00\% & 100.00\% & 95.83\% & 67.73\% & 46.99\% & 93.68\% & 78.06\% & 62.24\% & 77.18\% \\
Gemini-3-Flash & 97.78\% & 100.00\% & 34.29\% & 100.00\% & 100.00\% & 97.50\% & 97.96\% & 100.00\% & 95.83\% & 71.55\% & 43.37\% & 92.63\% & 77.55\% & 63.07\% & 81.21\% \\
Gemini-2.5-Pro & 100.00\% & 96.97\% & 23.43\% & 100.00\% & 96.30\% & 32.50\% & 61.22\% & 100.00\% & 97.92\% & 65.82\% & 39.76\% & 93.68\% & 72.96\% & 65.98\% & 73.49\% \\
Gemini-2.5-Flash & 97.78\% & 96.97\% & 28.00\% & 100.00\% & 85.19\% & 90.00\% & 97.96\% & 96.77\% & 95.83\% & 64.54\% & 42.17\% & 90.53\% & 79.08\% & 67.63\% & 78.19\% \\
Claude-Opus-4.5 & 100.00\% & 96.97\% & 44.57\% & 100.00\% & 100.00\% & 95.00\% & 100.00\% & 100.00\% & 97.92\% & 74.31\% & 63.86\% & 94.74\% & 83.16\% & 69.29\% & 73.15\% \\
Claude-Sonnet-4.5 & 100.00\% & 100.00\% & 28.57\% & 100.00\% & 96.30\% & 97.50\% & 100.00\% & 100.00\% & 95.83\% & 76.01\% & 54.22\% & 90.53\% & 73.98\% & 70.54\% & 74.83\% \\
Claude-Haiku-4.5 & 97.78\% & 96.97\% & 25.71\% & 100.00\% & 96.30\% & 95.00\% & 100.00\% & 100.00\% & 97.92\% & 67.30\% & 55.42\% & 93.68\% & 69.39\% & 61.41\% & 71.14\% \\
Grok-4.1 & 100.00\% & 90.91\% & 5.14\% & 100.00\% & 96.30\% & 95.00\% & 100.00\% & 100.00\% & 93.75\% & 73.67\% & 51.81\% & 92.63\% & 75.51\% & 67.22\% & 70.13\% \\
Grok-4-0709 & 95.56\% & 96.97\% & 19.43\% & 100.00\% & 92.59\% & 97.50\% & 100.00\% & 100.00\% & 93.75\% & 71.97\% & 51.81\% & 92.63\% & 73.47\% & 65.98\% & 73.15\% \\
\midrule
\multicolumn{16}{l}{\textbf{\textit{Open-source LLMs}}} \\
DeepSeek-V3.2 & 95.56\% & 96.97\% & 39.43\% & 94.74\% & 88.89\% & 95.00\% & 97.96\% & 100.00\% & 95.83\% & 60.30\% & 53.01\% & 91.58\% & 76.53\% & 68.88\% & 75.50\% \\
DeepSeek-R1 & 95.56\% & 96.97\% & 12.18\% & 100.00\% & 96.30\% & 95.00\% & 97.96\% & 100.00\% & 95.83\% & 70.49\% & 43.37\% & 89.47\% & 80.61\% & 68.05\% & 75.17\% \\
DeepSeek-R1-Distill-32B & 97.78\% & 90.91\% & 32.57\% & 100.00\% & 85.19\% & 90.00\% & 100.00\% & 100.00\% & 95.83\% & 66.88\% & 48.19\% & 94.74\% & 76.02\% & 72.61\% & 77.18\% \\
Qwen3-235B-Instruct & 97.78\% & 96.97\% & 42.29\% & 97.37\% & 81.48\% & 92.50\% & 97.96\% & 100.00\% & 93.75\% & 66.88\% & 48.19\% & 96.84\% & 81.63\% & 71.37\% & 74.50\% \\
Qwen2.5-72B-Instruct & 88.89\% & 100.00\% & 33.14\% & 100.00\% & 85.19\% & 97.50\% & 95.92\% & 96.77\% & 93.75\% & 64.12\% & 53.01\% & 90.53\% & 76.53\% & 73.44\% & 72.82\% \\
GLM-4.5 & 97.78\% & 96.97\% & 22.29\% & 100.00\% & 92.59\% & 92.50\% & 95.92\% & 96.77\% & 95.83\% & 69.00\% & 46.99\% & 92.63\% & 72.96\% & 70.12\% & 74.83\% \\
GLM-4 & 91.11\% & 90.91\% & 13.14\% & 92.11\% & 81.48\% & 80.00\% & 95.92\% & 93.55\% & 89.58\% & 54.35\% & 44.58\% & 85.26\% & 59.18\% & 64.32\% & 71.48\% \\
Llama-4-Scout & 93.33\% & 90.91\% & 15.43\% & 100.00\% & 88.89\% & 92.50\% & 100.00\% & 93.55\% & 95.83\% & 61.15\% & 37.35\% & 70.53\% & 62.24\% & 70.12\% & 73.15\% \\
Llama-3-70B-8192 & 88.89\% & 93.94\% & 46.29\% & 92.11\% & 70.37\% & 92.50\% & 97.96\% & 96.77\% & 93.75\% & 56.90\% & 45.78\% & 91.58\% & 69.39\% & 73.44\% & 79.87\% \\
Llama-3.3-70B-Instruct & 97.78\% & 93.94\% & 41.71\% & 97.37\% & 81.48\% & 92.50\% & 100.00\% & 93.55\% & 93.75\% & 60.72\% & 34.94\% & 92.63\% & 68.37\% & 73.86\% & 81.21\% \\
Llama-3.1-8B-Instruct & 88.89\% & 96.97\% & 47.70\% & 92.11\% & 74.07\% & 82.50\% & 97.96\% & 96.77\% & 87.50\% & 48.20\% & 36.14\% & 78.95\% & 60.20\% & 52.28\% & 76.85\% \\
Llama-3-8B-Instruct & 93.33\% & 100.00\% & 21.14\% & 100.00\% & 77.78\% & 92.50\% & 93.88\% & 96.77\% & 91.67\% & 52.23\% & 45.78\% & 90.53\% & 54.59\% & 61.41\% & 81.21\% \\
\bottomrule
\end{tabular}%
}
\caption{KE subsets include Factual Distortion (FD), Intra-Memory Conflict (IMC) and Entity-Identity Confusion (EIC). Abbreviations include Art=ArtCulture, Biz=Business, DefAn=DefAn, Food=FoodCooking, Lang=Language, LawCrimMil=LawCrimeMilitary, Rel=Religion, Sci=Science, Sports=Sports, Truth=TruthfulQA; SelfBuilt=SelfBuilt, WhoEnt=WhoQA\_Entity, WiC=WiC.}
\label{tab:prism-ke}
\end{table*}

% ========================= % KM % =========================
\begin{table*}[h]
\centering
\setlength{\tabcolsep}{4pt}
\renewcommand{\arraystretch}{0.9}
\resizebox{\textwidth}{!}{%
\begin{threeparttable}
\begin{tabular}{l cccccc cccccccccccc cc cc}
\toprule
\multirow{2}{*}{Model} & \multicolumn{6}{c}{DSK} & \multicolumn{12}{c}{FK} & \multicolumn{2}{c}{TK} & \multicolumn{2}{c}{NPK} \\
\cmidrule(lr){2-7}\cmidrule(lr){8-19}\cmidrule(lr){20-21}\cmidrule(lr){22-23}
& CnSafe & PubMed & RAG-J & RAG-F & Strategy & SciQ & Awards & Biology & Comp & Country & Dynasty & Festival & LLM & Literature & Military & Phone & Time & Univ & Numeric & Text & Org & Private \\
\midrule
\multicolumn{23}{l}{\textbf{\textit{Proprietary LLMs}}} \\
GPT-5.2 & 85.11\% & 74.03\% & 64.56\% & 88.68\% & 86.54\% & 87.40\% & 100.00\% & 100.00\% & 100.00\% & 100.00\% & 95.24\% & 100.00\% & 100.00\% & 100.00\% & 100.00\% & 100.00\% & 100.00\% & 100.00\% & 100.00\% & 99.08\% & 99.19\% & 97.57\% \\
GPT-5.1 & 84.57\% & 49.35\% & 37.97\% & 92.45\% & 51.92\% & 37.80\% & 100.00\% & 100.00\% & 100.00\% & 100.00\% & 95.24\% & 100.00\% & 100.00\% & 100.00\% & 100.00\% & 100.00\% & 100.00\% & 100.00\% & 100.00\% & 100.00\% & 100.00\% & 98.61\% \\
GPT-4o-20241120 & 81.91\% & 59.31\% & 49.37\% & 96.23\% & 73.08\% & 52.36\% & 100.00\% & 100.00\% & 100.00\% & 100.00\% & 95.24\% & 100.00\% & 100.00\% & 100.00\% & 100.00\% & 100.00\% & 100.00\% & 100.00\% & 99.40\% & 100.00\% & 98.79\% & 98.61\% \\
Gemini-3-Pro & 64.36\% & 58.01\% & 68.35\% & 94.34\% & 61.54\% & 25.98\% & 100.00\% & 98.98\% & 100.00\% & 100.00\% & 85.71\% & 88.46\% & 100.00\% & 100.00\% & 100.00\% & 100.00\% & 100.00\% & 93.75\% & 98.19\% & 95.41\% & 92.34\% & 93.06\% \\
Gemini-3-Flash & 82.98\% & 61.90\% & 64.56\% & 100.00\% & 73.08\% & 61.81\% & 100.00\% & 100.00\% & 100.00\% & 100.00\% & 100.00\% & 96.15\% & 100.00\% & 100.00\% & 100.00\% & 100.00\% & 100.00\% & 96.88\% & 98.80\% & 100.00\% & 93.15\% & 92.01\% \\
Gemini-2.5-Pro & 88.30\% & 61.90\% & 65.82\% & 92.45\% & 65.38\% & 50.00\% & 100.00\% & 100.00\% & 100.00\% & 100.00\% & 100.00\% & 100.00\% & 100.00\% & 100.00\% & 100.00\% & 100.00\% & 100.00\% & 100.00\% & 96.99\% & 100.00\% & 93.55\% & 87.85\% \\
Gemini-2.5-Flash & 80.32\% & 73.16\% & 60.76\% & 96.23\% & 73.08\% & 56.69\% & 100.00\% & 100.00\% & 100.00\% & 100.00\% & 90.48\% & 92.31\% & 96.55\% & 100.00\% & 100.00\% & 100.00\% & 100.00\% & 96.88\% & 98.19\% & 99.08\% & 98.79\% & 95.14\% \\
Claude-Opus-4.5 & 82.45\% & 85.28\% & 86.08\% & 98.11\% & 65.38\% & 82.28\% & 100.00\% & 89.80\% & 100.00\% & 100.00\% & 90.48\% & 100.00\% & 100.00\% & 100.00\% & 100.00\% & 100.00\% & 100.00\% & 96.88\% & 98.19\% & 92.66\% & 97.58\% & 95.14\% \\
Claude-Sonnet-4.5 & 86.70\% & 78.35\% & 65.82\% & 96.23\% & 71.15\% & 59.84\% & 100.00\% & 89.80\% & 100.00\% & 100.00\% & 100.00\% & 100.00\% & 100.00\% & 100.00\% & 100.00\% & 100.00\% & 100.00\% & 100.00\% & 100.00\% & 100.00\% & 98.39\% & 98.96\% \\
Claude-Haiku-4.5 & 89.89\% & 62.34\% & 56.96\% & 96.23\% & 71.15\% & 64.57\% & 100.00\% & 100.00\% & 100.00\% & 100.00\% & 100.00\% & 100.00\% & 100.00\% & 100.00\% & 100.00\% & 100.00\% & 100.00\% & 100.00\% & 100.00\% & 100.00\% & 100.00\% & 99.65\% \\
Grok-4.1 & 79.26\% & 58.87\% & 59.49\% & 37.74\% & 71.15\% & 34.25\% & 100.00\% & 97.96\% & 100.00\% & 100.00\% & 95.24\% & 100.00\% & 96.55\% & 79.25\% & 100.00\% & 100.00\% & 95.45\% & 96.88\% & 80.12\% & 51.38\% & 96.37\% & 95.49\% \\
Grok-4-0709 & 75.00\% & 67.53\% & 56.96\% & 56.60\% & 69.23\% & 30.31\% & 100.00\% & 98.98\% & 100.00\% & 100.00\% & 100.00\% & 100.00\% & 96.55\% & 86.79\% & 100.00\% & 100.00\% & 95.45\% & 96.88\% & 85.54\% & 71.56\% & 97.98\% & 94.10\% \\
\midrule
\multicolumn{23}{l}{\textbf{\textit{Open-source LLMs}}} \\
DeepSeek-V3.2 & 74.47\% & 62.77\% & 67.09\% & 94.34\% & 65.38\% & 69.69\% & 100.00\% & 100.00\% & 100.00\% & 100.00\% & 100.00\% & 100.00\% & 100.00\% & 100.00\% & 100.00\% & 100.00\% & 100.00\% & 93.75\% & 100.00\% & 100.00\% & 100.00\% & 97.92\% \\
DeepSeek-R1 & 77.66\% & 64.50\% & 62.03\% & 92.45\% & 65.38\% & 62.60\% & 100.00\% & 98.98\% & 100.00\% & 100.00\% & 95.24\% & 96.15\% & 100.00\% & 98.11\% & 100.00\% & 100.00\% & 95.45\% & 78.12\% & 99.40\% & 98.17\% & 98.79\% & 94.10\% \\
DeepSeek-R1-Distill-32B & 79.79\% & 59.74\% & 60.76\% & 84.91\% & 67.31\% & 59.84\% & 100.00\% & 97.96\% & 100.00\% & 95.65\% & 80.95\% & 96.15\% & 93.10\% & 96.23\% & 100.00\% & 100.00\% & 100.00\% & 90.62\% & 99.40\% & 97.25\% & 97.58\% & 94.79\% \\
Qwen3-235B-Instruct & 88.83\% & 70.13\% & 62.03\% & 98.11\% & 76.92\% & 51.18\% & 100.00\% & 100.00\% & 100.00\% & 100.00\% & 100.00\% & 100.00\% & 100.00\% & 100.00\% & 100.00\% & 100.00\% & 100.00\% & 100.00\% & 100.00\% & 100.00\% & 99.60\% & 99.31\% \\
Qwen2.5-72B-Instruct & 87.77\% & 62.77\% & 60.76\% & 96.23\% & 53.85\% & 49.21\% & 100.00\% & 100.00\% & 100.00\% & 100.00\% & 100.00\% & 100.00\% & 100.00\% & 100.00\% & 100.00\% & 100.00\% & 100.00\% & 100.00\% & 100.00\% & 100.00\% & 100.00\% & 100.00\% \\
GLM-4.5 & 73.40\% & 65.80\% & 65.82\% & 81.13\% & 63.46\% & 66.93\% & 100.00\% & 96.94\% & 100.00\% & 100.00\% & 85.71\% & 92.31\% & 100.00\% & 98.11\% & 100.00\% & 100.00\% & 95.45\% & 93.75\% & 98.80\% & 100.00\% & 94.35\% & 92.71\% \\
GLM-4 & 86.17\% & 54.98\% & 51.90\% & 96.23\% & 51.92\% & 43.31\% & 52.94\% & 97.96\% & 92.86\% & 100.00\% & 76.19\% & 96.15\% & 93.10\% & 81.13\% & 100.00\% & 100.00\% & 100.00\% & 71.88\% & 99.40\% & 100.00\% & 99.19\% & 98.96\% \\
Llama-4-Scout & 83.51\% & 70.13\% & 65.82\% & 96.23\% & 65.38\% & 58.27\% & 100.00\% & 100.00\% & 100.00\% & 100.00\% & 100.00\% & 100.00\% & 100.00\% & 100.00\% & 100.00\% & 100.00\% & 100.00\% & 96.88\% & 100.00\% & 100.00\% & 99.19\% & 97.92\% \\
Llama-3-70B-8192 & 83.51\% & 68.83\% & 69.62\% & 94.34\% & 67.31\% & 56.30\% & 100.00\% & 100.00\% & 100.00\% & 100.00\% & 100.00\% & 100.00\% & 100.00\% & 100.00\% & 100.00\% & 100.00\% & 100.00\% & 96.88\% & 100.00\% & 100.00\% & 100.00\% & 99.65\% \\
Llama-3.3-70B-Instruct & 84.57\% & 70.13\% & 78.48\% & 96.23\% & 71.15\% & 62.60\% & 100.00\% & 100.00\% & 100.00\% & 100.00\% & 100.00\% & 100.00\% & 100.00\% & 100.00\% & 100.00\% & 100.00\% & 100.00\% & 100.00\% & 100.00\% & 100.00\% & 100.00\% & 99.65\% \\
Llama-3.1-8B-Instruct & 52.13\% & 57.14\% & 62.03\% & 58.49\% & 51.92\% & 71.26\% & 97.06\% & 100.00\% & 100.00\% & 100.00\% & 80.95\% & 100.00\% & 82.76\% & 98.11\% & 100.00\% & 100.00\% & 100.00\% & 84.38\% & 100.00\% & 100.00\% & 98.79\% & 96.18\% \\
Llama-3-8B-Instruct & 34.04\% & 51.52\% & 58.23\% & 84.91\% & 55.77\% & 64.96\% & 88.24\% & 98.98\% & 85.71\% & 100.00\% & 80.95\% & 100.00\% & 89.66\% & 79.25\% & 100.00\% & 100.00\% & 90.91\% & 59.38\% & 98.80\% & 100.00\% & 91.94\% & 81.60\% \\
\bottomrule
\end{tabular}%
% \begin{tablenotes}
%         \item 100.00\%~indicates the model achieved 100\% accuracy.
%       \end{tablenotes}
    \end{threeparttable}

}

\caption{KM subsets include Domain-Specific (DSK), Fictional (FK), Timely (TK), and Non-Public (NPK) knowledge. Abbreviations include CnSafe=Chinese SafetyQA; PubMed=PubMedQA; RAG-J/F=RAG-QA-Leaderboard (Judge/Fill); Strategy=StrategyQA; Comp=Competition; Univ=University; Org=Organizational.}
\label{tab:prism-km}
\end{table*}

% ========================= % RE % =========================
\begin{table*}[h]
\centering
\setlength{\tabcolsep}{4pt}
\renewcommand{\arraystretch}{0.9}
\resizebox{\textwidth}{!}{%
    \begin{threeparttable}
\begin{tabular}{l ccc cc ccc cc}
\toprule
\multirow{2}{*}{Model} & \multicolumn{3}{c}{LF} & \multicolumn{2}{c}{PF} & \multicolumn{3}{c}{IIF} & \multicolumn{2}{c}{MRF} \\
\cmidrule(lr){2-4}\cmidrule(lr){5-6}\cmidrule(lr){7-9}\cmidrule(lr){10-11}
& CR& CI& CT& SubQ & Triple & Sum & Simpl & Dial & Math & Code \\
\midrule
\multicolumn{11}{l}{\textbf{\textit{Proprietary LLMs}}} \\
GPT-5.2 & 81.55\% & 85.31\% & 88.89\% & 63.66\% & 97.44\% & 89.28\% & 93.37\% & 85.60\% & 37.39\% & 1.54 \\
GPT-5.1 & 82.40\% & 86.36\% & 87.18\% & 72.97\% & 100.00\% & 85.78\% & 89.74\% & 82.84\% & 24.15\% & 1.42 \\
GPT-4o-20241120 & 76.82\% & 78.32\% & 79.49\% & 75.68\% & 97.44\% & 83.86\% & 92.63\% & 86.17\% & 11.81\% & 1.03 \\
Gemini-3-Pro & 84.98\% & 88.11\% & 92.31\% & 86.79\% & 100.00\% & 88.81\% & 96.47\% & 84.44\% & 57.25\% & 2.45 \\
Gemini-3-Flash & 87.12\% & 88.46\% & 95.73\% & 75.68\% & 100.00\% & 85.84\% & 92.25\% & 75.35\% & 52.06\% & 1.58 \\
Gemini-2.5-Pro & 87.98\% & 89.51\% & 93.16\% & 72.37\% & 100.00\% & 88.14\% & 92.78\% & 83.22\% & 42.93\% & 1.68 \\
Gemini-2.5-Flash & 70.82\% & 75.87\% & 67.52\% & 55.26\% & 98.29\% & 84.23\% & 93.65\% & 84.95\% & 30.23\% & 1.17 \\
Claude-Opus-4.5 & 87.12\% & 87.76\% & 90.60\% & 66.37\% & 99.15\% & 85.84\% & 90.48\% & 86.68\% & 53.49\% & 2.79 \\
Claude-Sonnet-4.5 & 84.55\% & 87.41\% & 85.47\% & 66.07\% & 98.29\% & 87.51\% & 91.32\% & 79.90\% & 39.89\% & 2.21 \\
Claude-Haiku-4.5 & 76.39\% & 81.47\% & 83.76\% & 50.75\% & 100.00\% & 82.58\% & 89.56\% & 81.96\% & 43.47\% & 2.71 \\
Grok-4.1 & 73.82\% & 79.37\% & 78.63\% & 69.67\% & 42.74\% & 85.45\% & 86.73\% & 77.52\% & 28.09\% & 1.43 \\
Grok-4-0709 & 79.40\% & 85.31\% & 83.76\% & 69.97\% & 61.54\% & 85.38\% & 87.68\% & 78.51\% & 28.62\% & 1.51 \\
\midrule
\multicolumn{11}{l}{\textbf{\textit{Open-source LLMs}}} \\
DeepSeek-V3.2 & 74.68\% & 74.83\% & 78.63\% & 71.77\% & 99.15\% & 85.74\% & 89.14\% & 87.09\% & 30.05\% & 1.29 \\
DeepSeek-R1 & 82.40\% & 87.06\% & 90.60\% & 72.97\% & 94.02\% & 86.23\% & 92.73\% & 79.92\% & 21.65\% & 0.60 \\
DeepSeek-R1-Distill-32B & 83.69\% & 86.71\% & 89.74\% & 63.36\% & 100.00\% & 82.66\% & 91.85\% & 77.38\% & 13.42\% & 0.36 \\
Qwen3-235B-Instruct & 85.41\% & 88.11\% & 84.62\% & 72.37\% & 100.00\% & 84.61\% & 91.68\% & 79.73\% & 22.54\% & 1.60 \\
Qwen2.5-72B-Instruct & 81.97\% & 82.52\% & 83.76\% & 72.67\% & 98.29\% & 82.15\% & 89.80\% & 83.90\% & 12.52\% & 1.78 \\
GLM-4.5 & 80.69\% & 86.01\% & 82.05\% & 64.56\% & 96.58\% & 87.04\% & 91.96\% & 80.20\% & 8.05\% & 1.47 \\
GLM-4 & 59.23\% & 66.08\% & 58.12\% & 76.88\% & 98.29\% & 83.98\% & 87.72\% & 87.72\% & 4.47\% & 1.09 \\
Llama-4-Scout & 68.67\% & 69.23\% & 69.23\% & 67.57\% & 95.73\% & 85.78\% & 87.31\% & 80.59\% & 18.43\% & 1.78 \\
Llama-3-70B-8192 & 75.97\% & 77.97\% & 78.63\% & 68.77\% & 23.08\% & 85.42\% & 90.86\% & 83.44\% & 11.81\% & 1.45 \\
Llama-3.3-70B-Instruct & 76.82\% & 81.82\% & 81.20\% & 64.26\% & 99.15\% & 85.42\% & 90.03\% & 82.12\% & 11.81\% & 1.63 \\
Llama-3.1-8B-Instruct & 33.48\% & 27.97\% & 33.33\% & 60.36\% & 25.64\% & 82.70\% & 84.50\% & 84.20\% & 4.83\% & 0.90 \\
Llama-3-8B-Instruct & 34.76\% & 35.31\% & 35.04\% & 53.75\% & 29.91\% & 79.99\% & 83.95\% & 84.29\% & 5.37\% & 0.87 \\
\bottomrule
\end{tabular}%
% \begin{tablenotes}
%         \item 100.00\%~indicates the model achieved 100\% accuracy.
%       \end{tablenotes}
    \end{threeparttable}
}
\caption{RE subsets include Logical Fallacy(LF), Procedural Failure(PF), Information Integration Failure(IIF), Mathematical Reasoning Failure(MRF). Abbreviations include CR=Critical Reasoning; CI=Contextual Inference; CT=Cognitive Traps; SubQ = Sub-question Decomposition; Triple = Triplet-based Hopping; Sum = Summarization; Simpl=Simplification; Dial = Dialogue Extraction; Math = Mathematics; Code = Code Generation.}
\label{tab:prism-re}
\end{table*}

% ========================= % IFE % =========================
\begin{table*}[!h]
\centering
\setlength{\tabcolsep}{4pt}
\renewcommand{\arraystretch}{0.9}
\resizebox{\textwidth}{!}{%
    \begin{threeparttable}

\begin{tabular}{l cc ccc cccccc ccccccc}
\toprule
\multirow{2}{*}{Model} & \multicolumn{2}{c}{EF} & \multicolumn{3}{c}{LC} & \multicolumn{6}{c}{LgC} & \multicolumn{7}{c}{CCL} \\
\cmidrule(lr){2-3}\cmidrule(lr){4-6}\cmidrule(lr){7-12}\cmidrule(lr){13-19}
& Schema & TextStruct & Approx & LB & UB & DE & ES & FR & JA & RU & ZH & Forbid & KeyIncl & LenMix & Math & NoComma & StartEnd & TwoResp \\
\midrule
\multicolumn{19}{l}{\textbf{\textit{Proprietary LLMs}}} \\
GPT-5.2 & 87.04\% & 36.54\% & 55.95\% & 75.00\% & 94.26\% & 92.79\% & 88.89\% & 87.74\% & 93.55\% & 90.00\% & 81.42\% & 93.88\% & 96.49\% & 82.52\% & 83.00\% & 93.94\% & 100.00\% & 90.77\% \\
GPT-5.1 & 54.30\% & 47.77\% & 45.95\% & 78.23\% & 59.84\% & 90.99\% & 90.11\% & 92.45\% & 87.10\% & 92.00\% & 73.45\% & 75.51\% & 87.72\% & 62.24\% & 78.33\% & 74.24\% & 68.66\% & 80.00\% \\
GPT-4o-20241120 & 86.09\% & 37.58\% & 60.27\% & 62.90\% & 96.72\% & 92.79\% & 94.51\% & 94.34\% & 93.55\% & 94.00\% & 81.42\% & 91.84\% & 89.47\% & 66.43\% & 81.67\% & 87.88\% & 91.04\% & 95.38\% \\
Gemini-3-Pro & 66.42\% & 34.39\% & 96.50\% & 100.00\% & 99.17\% & 97.27\% & 95.56\% & 96.04\% & 93.48\% & 95.92\% & 78.18\% & 97.96\% & 98.13\% & 80.15\% & 86.31\% & 95.24\% & 93.94\% & 100.00\% \\
Gemini-3-Flash & 45.21\% & 33.76\% & 81.62\% & 74.19\% & 98.36\% & 96.40\% & 96.70\% & 94.34\% & 96.77\% & 94.95\% & 79.46\% & 93.88\% & 94.74\% & 66.67\% & 89.33\% & 96.97\% & 80.30\% & 98.44\% \\
Gemini-2.5-Pro & 62.33\% & 33.12\% & 71.35\% & 95.16\% & 86.89\% & 97.30\% & 96.70\% & 96.23\% & 94.62\% & 95.96\% & 82.14\% & 93.88\% & 99.12\% & 78.72\% & 87.00\% & 95.45\% & 91.04\% & 93.85\% \\
Gemini-2.5-Flash & 58.22\% & 35.03\% & 59.46\% & 94.35\% & 89.34\% & 96.40\% & 97.80\% & 91.51\% & 89.25\% & 93.94\% & 77.68\% & 93.88\% & 99.12\% & 75.89\% & 85.67\% & 98.48\% & 94.03\% & 90.77\% \\
Claude-Opus-4.5 & 55.03\% & 36.31\% & 59.46\% & 99.19\% & 72.95\% & 96.40\% & 95.60\% & 94.34\% & 96.77\% & 96.00\% & 82.30\% & 89.80\% & 97.37\% & 72.03\% & 81.67\% & 95.45\% & 97.01\% & 98.46\% \\
Claude-Sonnet-4.5 & 27.81\% & 35.67\% & 64.59\% & 100.00\% & 69.67\% & 95.50\% & 92.31\% & 96.23\% & 97.85\% & 98.00\% & 83.19\% & 95.92\% & 96.49\% & 70.63\% & 87.67\% & 90.91\% & 98.51\% & 95.38\% \\
Claude-Haiku-4.5 & 25.83\% & 33.76\% & 52.70\% & 96.77\% & 72.13\% & 98.20\% & 97.80\% & 95.28\% & 96.77\% & 96.00\% & 85.84\% & 89.80\% & 92.98\% & 62.94\% & 91.67\% & 95.45\% & 94.03\% & 95.38\% \\
Grok-4.1 & 65.75\% & 38.06\% & 45.95\% & 100.00\% & 82.79\% & 95.50\% & 92.31\% & 94.34\% & 93.55\% & 97.00\% & 78.76\% & 81.63\% & 93.86\% & 65.03\% & 83.33\% & 87.88\% & 82.09\% & 61.54\% \\
Grok-4-0709 & 86.00\% & 35.03\% & 43.24\% & 99.19\% & 81.15\% & 94.59\% & 95.60\% & 94.34\% & 95.70\% & 95.00\% & 84.07\% & 91.84\% & 97.37\% & 71.33\% & 84.01\% & 90.91\% & 89.55\% & 87.69\% \\
\midrule
\multicolumn{19}{l}{\textbf{\textit{Open-source LLMs}}} \\
DeepSeek-V3.2 & 77.48\% & 32.48\% & 47.57\% & 99.19\% & 86.07\% & 96.40\% & 94.51\% & 96.23\% & 90.32\% & 97.00\% & 84.96\% & 95.92\% & 96.49\% & 69.93\% & 90.00\% & 96.97\% & 94.03\% & 96.92\% \\
DeepSeek-R1 & 75.57\% & 35.29\% & 84.91\% & 98.17\% & 95.65\% & 93.69\% & 91.21\% & 93.40\% & 92.47\% & 97.98\% & 82.30\% & 97.83\% & 97.00\% & 80.00\% & 83.16\% & 96.83\% & 92.31\% & 96.83\% \\
DeepSeek-R1-Distill-32B & 69.72\% & 37.58\% & 49.45\% & 91.94\% & 80.99\% & 93.69\% & 89.01\% & 92.45\% & 93.55\% & 94.00\% & 80.53\% & 89.80\% & 91.89\% & 71.63\% & 82.95\% & 98.48\% & 94.03\% & 61.54\% \\
Qwen3-235B-Instruct & 74.48\% & 38.22\% & 42.70\% & 99.19\% & 60.66\% & 92.79\% & 86.81\% & 85.85\% & 89.25\% & 91.92\% & 83.93\% & 91.84\% & 92.98\% & 79.43\% & 94.00\% & 95.45\% & 89.55\% & 89.23\% \\
Qwen2.5-72B-Instruct & 85.42\% & 36.94\% & 30.54\% & 67.74\% & 98.36\% & 97.30\% & 93.41\% & 91.51\% & 92.47\% & 93.94\% & 81.25\% & 95.92\% & 89.47\% & 72.14\% & 82.67\% & 100.00\% & 76.12\% & 83.08\% \\
GLM-4.5 & 76.11\% & 32.90\% & 61.41\% & 87.90\% & 94.26\% & 94.44\% & 96.67\% & 90.48\% & 92.47\% & 94.00\% & 77.68\% & 91.67\% & 94.74\% & 73.94\% & 86.99\% & 96.88\% & 89.55\% & 96.88\% \\
GLM-4 & 55.32\% & 35.67\% & 24.32\% & 66.13\% & 95.08\% & 93.69\% & 89.01\% & 87.74\% & 92.47\% & 88.00\% & 68.14\% & 75.51\% & 74.56\% & 63.38\% & 83.33\% & 77.27\% & 85.07\% & 89.23\% \\
Llama-4-Scout & 74.29\% & 38.46\% & 65.95\% & 92.74\% & 98.36\% & 99.10\% & 98.90\% & 97.17\% & 93.55\% & 96.00\% & 82.30\% & 95.56\% & 93.86\% & 81.12\% & 84.33\% & 98.48\% & 67.16\% & 87.69\% \\
Llama-3-70B-8192 & 76.87\% & 35.67\% & 55.14\% & 70.97\% & 100.00\% & 97.30\% & 97.80\% & 92.45\% & 88.17\% & 95.00\% & 84.07\% & 91.84\% & 92.98\% & 72.73\% & 89.67\% & 95.45\% & 82.09\% & 96.92\% \\
Llama-3.3-70B-Instruct & 90.78\% & 35.90\% & 37.84\% & 83.06\% & 100.00\% & 99.10\% & 98.90\% & 96.23\% & 95.70\% & 97.00\% & 84.96\% & 95.92\% & 95.61\% & 88.81\% & 88.00\% & 96.97\% & 86.57\% & 92.31\% \\
Llama-3.1-8B-Instruct & 49.37\% & 31.85\% & 38.38\% & 33.87\% & 97.54\% & 98.20\% & 95.60\% & 95.28\% & 93.55\% & 93.00\% & 74.34\% & 95.92\% & 81.58\% & 58.04\% & 89.86\% & 90.91\% & 80.60\% & 81.54\% \\
Llama-3-8B-Instruct & 38.51\% & 29.30\% & 24.59\% & 23.39\% & 100.00\% & 95.50\% & 90.11\% & 89.62\% & 47.31\% & 91.00\% & 40.71\% & 83.67\% & 83.33\% & 62.94\% & 84.67\% & 89.39\% & 68.18\% & 72.31\% \\
\bottomrule
\end{tabular}%
% \begin{tablenotes}
%         \item 100.00\%~indicates the model achieved 100\% accuracy.
%       \end{tablenotes}
    \end{threeparttable}
}
\caption{IFE subsets include Explicit Format(EF), Length Constraints(LC), Language Constraints(LgC), and Complex \& Cognitive Load(CCL). Abbreviations include Schema=Data Schema; TextStruct=Text Structure; Approx=Approximate; LB=Lower Bound; UB=Upper Bound; DE/ES/FR/JA/RU/ZH=German/Spanish/French/Japanese/Russian/Chinese; Forbid=Forbidden Words; KeyIncl=Keyword Inclusion; LenMix=Length/Format Mixed; Math=Math Reasoning; NoComma=No Comma; StartEnd=Start/End Phrase; TwoResp=Two Responses.}
\label{tab:prism-ife}
\end{table*}

\section{Sampling Parameters Grid Search}
\label{app:decoding_grid}

We search temperature in $\{0,0.2,0.4,0.6,0.8\}$ and top-$p$ in $\{0.6,0.8,0.9,0.95\}$.
For each dimension, we choose the configuration that yields the lowest hallucination rate on the development split.
Tables~\ref{tab:hyper1} and~\ref{tab:hyper2} report the full grid results.

For the LLM-Eval judge, we use the same grid and repeat judging 10 times with different random seeds.
Table~\ref{tab:hyper3} reports the mean and standard deviation of the judge scores under each configuration.

\begin{table*}[h]
\centering
\resizebox{\linewidth}{!}{
\begin{tabular}{c|ccccc|ccccc}
\toprule
\multicolumn{6}{c|}{KM (Metric: Hallucination Rate)} & \multicolumn{5}{c}{KE (Metric: Hallucination Rate)} \\
\midrule
& \text{Temp.=0} & \text{Temp.=0.2} & \text{Temp.=0.4} & \text{Temp.=0.6} & \text{Temp.=0.8} &\text{Temp.=0} & \text{Temp.=0.2} & \text{Temp.=0.4} & \text{Temp.=0.6} & \text{Temp.=0.8} \\
\midrule
Top-p=0.6 & \multirow{4}{*}{34.78\%} &34.78\% &34.78\% &34.78\% &26.09\% & \multirow{4}{*}{44.90\%} &40.82\% &42.86\% &44.90\% &44.90\% \\
Top-p=0.8 & &34.78\% &26.09\% &34.78\% &\cellcolor{rank1}26.09\% & &44.90\% &38.78\% &42.86\% &\cellcolor{rank1}36.73\% \\
Top-p=0.9& &34.78\% &30.43\% &39.13\% &26.09\% & &48.98\% &48.98\% &42.86\% &53.06\% \\
Top-p=0.95& &30.43\% &30.43\% &39.13\% &30.43\% & &46.94\% &36.73\% &51.02\% &53.06\% \\
\bottomrule
\end{tabular}
}
\caption{Hyperparameter grid search results. The selected hyperparameter combinations have been highlighted.}
\label{tab:hyper1}
\end{table*}

\begin{table*}[h]
\centering
\resizebox{\linewidth}{!}{
\begin{tabular}{c|ccccc|ccccc}
\toprule
\multicolumn{6}{c|}{RE (Metric: Hallucination Rate)} & \multicolumn{5}{c}{IFE (Metric: Hallucination Rate)} \\
\midrule
& \text{Temp.=0} & \text{Temp.=0.2} & \text{Temp.=0.4} & \text{Temp.=0.6} & \text{Temp.=0.8} &\text{Temp.=0} & \text{Temp.=0.2} & \text{Temp.=0.4} & \text{Temp.=0.6} & \text{Temp.=0.8} \\
\midrule
Top-p=0.6 & \multirow{4}{*}{78.79\%} &74.24\% &77.27\% &72.73\% &77.27\% & \multirow{4}{*}{————} &5.26\% &5.26\% &10.53\% &5.26\% \\
Top-p=0.8 & &74.24\% &77.27\% &71.21\% &83.33\% & &10.53\% &10.53\% &15.79\% &\cellcolor{rank1}3.67\% \\
Top-p=0.9& &75.76\% &81.82\% &77.27\% &71.21\% & &5.26\% &10.53\% &10.53\% &10.53\% \\
Top-p=0.95& &83.33\% &\cellcolor{rank1}66.67\% &72.73\% &74.24\% & &10.53\% &10.53\% &5.26\% &21.05\% \\
\bottomrule
\end{tabular}
}
\caption{Hyperparameter grid search results. The selected hyperparameter combinations have been highlighted.}
\label{tab:hyper2}
\end{table*}

\begin{table*}[!ht]
\centering
\begin{tabular}{c|ccccc}
\toprule
\multicolumn{6}{c}{RE (Metric: LLM-Eval)} \\
\midrule
& \text{Temp.=0} & \text{Temp.=0.2} & \text{Temp.=0.4} & \text{Temp.=0.6} & \text{Temp.=0.8} \\
\midrule
Top-p=0.6  & $1.28_{\pm0.03}$ & $1.30_{\pm0.05}$ & $1.29_{\pm0.07}$ & $1.33_{\pm0.09}$ & $1.35_{\pm0.17}$ \\
Top-p=0.8  & $1.42_{\pm0.04}$ & $1.46_{\pm0.06}$ & $1.51_{\pm0.08}$ & $1.55_{\pm0.10}$ & \cellcolor{rank1}$1.62_{\pm0.11}$ \\
Top-p=0.9  & $1.34_{\pm0.04}$ & $1.36_{\pm0.07}$ & $1.38_{\pm0.10}$ & $1.40_{\pm0.13}$ & $1.41_{\pm0.18}$ \\
Top-p=0.95 & $1.21_{\pm0.05}$ & $1.23_{\pm0.08}$ & $1.25_{\pm0.11}$ & $1.26_{\pm0.16}$ & $1.27_{\pm0.22}$ \\
\bottomrule
\end{tabular}
\caption{Hyperparameter search results for LLM-Eval. The selected hyperparameter combinations have been highlighted.}
\label{tab:hyper3}
\end{table*}

\begin{table}[ht] 
\centering
\small
\setlength{\tabcolsep}{5pt}
\begin{tabular}{l ccc c cc}
\toprule
& \multicolumn{4}{c}{\textbf{Metrics}} & \multicolumn{2}{c}{\textbf{Selection}} \\
\cmidrule(lr){2-5} \cmidrule(lr){6-7}
Domain & $C$ & $G$ & $V$ & $L$ & $S$ & Rank \\
\midrule
Festival     & 0.22 & 0.78 & 0.55 & 0.86 & \textbf{0.727} & 1 \\
LLMs         & 0.26 & 0.74 & 0.91 & 0.38 & 0.686 & 2 \\
Dynasty      & 0.24 & 0.76 & 0.48 & 0.81 & 0.681 & 3 \\
Awards       & 0.28 & 0.72 & 0.83 & 0.46 & 0.677 & 4 \\
Competition  & 0.31 & 0.69 & 0.79 & 0.40 & 0.634 & 5 \\
Literature   & 0.29 & 0.71 & 0.44 & 0.74 & 0.628 & 6 \\
Phone        & 0.33 & 0.67 & 0.88 & 0.29 & 0.623 & 7 \\
Time         & 0.35 & 0.65 & 0.76 & 0.31 & 0.582 & 8 \\
Biology      & 0.17 & 0.83 & 0.52 & 0.34 & 0.574 & 9 \\
Military     & 0.21 & 0.79 & 0.57 & 0.33 & 0.574 & 10 \\
University   & 0.53 & 0.47 & 0.49 & 0.27 & 0.415 & 11 \\
Country      & 0.68 & 0.32 & 0.41 & 0.22 & 0.319 & 12 \\
\bottomrule
\end{tabular}
\caption{Selected domains for KM\_FK ranked by the combined score $S$. The metrics are Exposure ($C$), Gap ($G$), Speed ($V$), and Language Spread ($L$).}
\label{tab:kmfk_domains}
\end{table}

% \section{Further Analysis}
% \input{src/none_Interpretability}

\section{More Results}
\subsection{Language Selection Rationale for IFE\_LgC}
We selected five high-resource target languages: ZH, FR, DE, RU, and ES. This selection is grounded in their core status within web-scale corpus and mainstream LLM training mixtures. According to the analysis of the CCNet corpus by \citet{wenzek2019ccnet}, these five languages consistently occupy the top clusters of high-quality web data.When English is excluded, RU (5.9\%), DE (5.8\%), and ZH (5.6\%) rank 2nd, 3rd, and 4th globally, followed closely by FR (3.9\%) and ES (3.7\%). Together, they account for over 24.9\% of the total web corpus, far exceeding the sum of all other long-tail languages. This dominance persists in both commercial and open-science models. Even in GPT-3 \citep{brown2020language}, which exhibits a strong Anglocentric bias (92.6\% English), FR (1.8\%), DE (1.5\%), and ES (0.8\%) remain the primary auxiliary knowledge sources. Similarly, the ROOTS corpus constructed for the BLOOM model \citep{laurencon2022roots} explicitly elevates ZH (16.2\%), FR (12.9\%), and ES (10.8\%) to the top three positions after English (30.04\%), establishing them as standard benchmarks for evaluating cross-lingual generalization.

\subsection{Domain Selection Rationale for KM\_FK}

KM\_FK targets mistakes caused by missing facts.
Prior work demonstrates that missing subject facts can lead to erroneous answers, and that multiple conditions can trigger the mixing of facts \citep{yu-etal-2024-mechanistic,zhang2024knowledge,wu2025enhancing}.
To ensure the domain choice is fair and reproducible, we select domains via a systematic scoring procedure.

\paragraph{Candidate Domains.}
We construct a candidate list from a public knowledge base and use the Wikidata Query Service to collect a set of items $E_d$ for each domain $d$ \citep{wikidatawdqs}.
This approach fixes the domain boundary and supports repeatable data collection.

\paragraph{Domain Scores.}
For each domain, we compute three scores and scale each to the range $[0, 1]$ over the full candidate list.
A larger value indicates a higher risk of missing knowledge.

\begin{itemize}[leftmargin=*, nosep, itemsep=4pt]
    \item \textbf{Exposure Gap:} 
    We estimate the exposure $C(d)$ by matching items in $E_d$ against public large-scale corpora widely used as training sources, including C4, The Pile, and ROOTS \citep{raffel2020t5,gao2020pile}.
    We define the gap as:
    \begin{equation}
        G(d) = 1 - C(d).
    \end{equation}

    \item \textbf{Update Speed:} 
    We measure how frequently facts change by analyzing recent edits of items in $E_d$ from Wikidata.
    Let $u(e)$ be the number of edits for item $e$ within a fixed time window.
    We define:
    \begin{equation}
        V(d) = \mathrm{scale}\!\left(\frac{1}{|E_d|}\sum_{e\in E_d} u(e)\right).
    \end{equation}

    \item \textbf{Language Spread:} 
    We measure the unevenness of coverage across languages.
    Let $p_{d,\ell}$ be the proportion of items in $E_d$ having labels in language $\ell$ among the top $m$ languages.
    We define:
    \begin{equation}
        L(d) = 1 - \frac{-\sum_{\ell=1}^{m} p_{d,\ell}\log p_{d,\ell}}{\log m}.
    \end{equation}
    This metric is motivated by known quality and coverage disparities in web-crawled multilingual data \citep{xue2021mt5, kreutzer-etal-2022-quality}.
\end{itemize}

\paragraph{Objective Weights.}
We combine the three scores as:
\begin{equation}
    S(d) = w_G G(d) + w_V V(d) + w_L L(d),
\end{equation}
subject to $w_G+w_V+w_L=1$.
We verify weights from the candidate list using the CRITIC method \citep{diakoulaki1995critic}.
Let $x_{ij}$ be the value of score $j$ for domain $i$, where $j\in\{G,V,L\}$ and there are $n$ candidate domains.
We first apply min-max scaling:
\begin{equation}
    z_{ij} = \frac{x_{ij}-\min_i x_{ij}}{\max_i x_{ij}-\min_i x_{ij}}.
\end{equation}
We then compute the standard deviation $\sigma_j$ and the correlation $r_{jk}$ between score $j$ and score $k$.
CRITIC defines the information quantity $c_j$ and weight $w_j$ as:
\begin{equation}
    c_j = \sigma_j\sum_{k\neq j}(1-r_{jk}), \quad
    w_j = \frac{c_j}{\sum_{t\in\{G,V,L\}} c_t}.
\end{equation}
In our setting, this yields $w_G=0.354$, $w_V=0.337$, and $w_L=0.309$.

\paragraph{Selection.}
We compute $S(d)$ for all candidate domains and select the top 12 domains for KM\_FK.
Table~\ref{tab:kmfk_domains} reports the detailed scores.

\section{Cases of KM}
\label{km}

The model faces two competing sources of truth: (1) the prompt text asserting “Paris, France” and (2) its internal world knowledge that Einstein was born in Ulm. Because instruction-following models prioritize immediate context, the passage is treated as authoritative, leading the model to answer “Paris, France” while sometimes adding a caveat about “Ulm, Germany.” Absent an explicit verification requirement or external tool, the model does not resolve which claim is correct and instead aligns with the most locally salient evidence, allowing prompt content to override memory and produce a confident but incorrect real-world answer.

\section{LLM Evaluation Details}
\label{llm_evaluation}
\subsection{Consistency with Human Annotations}
To validate the reliability of our \textit{LLM-as-a-judge} evaluation pipeline, we randomly sampled 100 questions from the RE\_code subset of \textbf{\textit{PRISM}} and asked five domain experts to score them using the same prompt and rubric as GPT-4o. The experts’ mean score is treated as the human reference, and we compute the absolute difference from GPT-4o’s score for each item, reporting the mean absolute deviation (MAD) and standard deviation (SD). The results indicate close agreement, with a MAD of 0.07 and an SD of 0.09.

\begin{ReasonBox}{Chain of Thought at Qwen3-4B}
{
</think> The original text states: ``Albert Einstein was born in Paris, France in 1879...'' So according to that, the answer is Paris, France. But wait, I remember that Einstein was actually born in Ulm, Germany. Hmm, maybe there's a mistake in the initial information given? Or perhaps the user is testing if I can confirm the correct answer based on the provided text. Wait, the user is asking based on the information given in the initial text. The initial text says he was born in Paris, France. But in reality, he was born in Ulm, Germany. However, since the user provided that specific information, I should stick to that. Unless the user is expecting the correct historical fact. But the question is phrased as ``Where was Albert Einstein born?'' and the initial text says Paris, France. So the answer should be Paris, France as per the given information. However, I need to check if there's any possible confusion here. Maybe the user is confused between the actual birthplace and the information given. But since the user provided the text, the answer should be based on that. Therefore, the answer is Paris, France. But I should also note that in reality, he was born in Ulm, Germany. But the question is based on the given text, so I should follow that. </think>
}
\end{ReasonBox}
% The model has two competing sources of truth:
% (1) the explicit prompt text claiming “Paris, France” and (2) its stored world knowledge that Einstein was born in Ulm. 
% Because instruction-following models are optimized to prioritize immediate context, they treat the passage as authoritative for the task, then resolve the conflict by answering “Paris, France” (consistent with the prompt) while optionally adding a caveat about “Ulm, Germany.”
% Without an explicit requirement or tool to verify facts externally, the model does not truly arbitrate which claim is correct; it aligns to the most locally salient evidence (the quoted text), which can override memory and yield a confident-but-wrong real-world answer.

\subsection{Prompt for LLM Eval}

\onecolumn

\vspace{-5pt}
\begin{figure*}[!h]
\centering
\begin{tcolorbox}[
  colback=aclback,
  colframe=aclbluer,
  coltitle=white,
  title=\textbf{Prompt for RE\_MRF\_Code Eval},
  % Hurry up, it’s 1.5ddl
  fonttitle=\large\bfseries,
  boxrule=1.2pt,
  arc=3mm,
  width=1\textwidth,
  left=6pt, right=6pt, top=4pt, bottom=4pt
]
\textbf{Role:} Code evaluation expert.\\
\textbf{Task:} Given a problem, test input/output, and a model-generated solution, score the solution on four dimensions (0--10) and output a JSON object.\\

\vspace{4pt}
\textbf{Scoring (0--10):}
\begin{itemize}[leftmargin=1.2em, itemsep=2pt, topsep=2pt]
  \item \textbf{Correctness:} Logic validity and whether the solution produces the expected output.
  \begin{itemize}[leftmargin=1.2em, itemsep=1pt, topsep=1pt]
    \item 10: Fully correct; passes all tests.
    \item 7--9: Core logic correct; minor bugs or edge-case issues.
    \item 4--6: Approach is plausible but implementation has major issues; fails some tests.
    \item 1--3: Fundamental logic errors; largely fails to solve the task.
    \item 0: Completely wrong or no meaningful code.
  \end{itemize}

  \item \textbf{Completeness:} Executability and whether key components (definitions, imports, I/O handling) are present.
  \begin{itemize}[leftmargin=1.2em, itemsep=1pt, topsep=1pt]
    \item 10: Complete and runnable as-is.
    \item 7--9: Mostly complete; small fixes needed to run.
    \item 4--6: Missing critical parts; not directly runnable.
    \item 1--3: Fragment/pseudocode; many essential parts missing.
    \item 0: No code.
  \end{itemize}

  \item \textbf{Efficiency:} Time/space complexity relative to expected constraints.
  \begin{itemize}[leftmargin=1.2em, itemsep=1pt, topsep=1pt]
    \item 10: Optimal or near-optimal complexity.
    \item 7--9: Efficient enough for typical constraints.
    \item 4--6: Works but has noticeable inefficiencies.
    \item 1--3: Very inefficient; likely to time out or exceed memory.
    \item 0: Not assessable (no code or insufficient information).
  \end{itemize}

  \item \textbf{Readability:} Clarity of structure, naming, modularity, and comments.
  \begin{itemize}[leftmargin=1.2em, itemsep=1pt, topsep=1pt]
    \item 10: Very clear and well-structured.
    \item 7--9: Generally readable with minor issues.
    \item 4--6: Understandable but poorly organized or unclear in places.
    \item 1--3: Hard to follow.
    \item 0: Not assessable.
  \end{itemize}
\end{itemize}

\vspace{4pt}
\textbf{Output Format:}
\begin{quote}\small
\{"correctness":0-10,"completeness":0-10,"efficiency":0-10,"readability":0-10,"total":0-5\}
\end{quote}
\textbf{Total:} \(\mathrm{round}\left(\frac{\text{correctness}+\text{completeness}+\text{efficiency}+\text{readability}}{40}\times 5\right)\).

\vspace{4pt}
\textbf{Example:}
\begin{quote}\small
\{"correctness":8,"completeness":9,"efficiency":7,"readability":8,"total":4\}
\end{quote}
\end{tcolorbox}
\vspace{-5pt}
\label{fig:codeeval_4d}
\end{figure*}

\begin{figure*}[!h]
\centering

% =======================
% RE_IIF_Sum
% =======================
\begin{tcolorbox}[
  colback=aclback,
  colframe=aclbluer,
  coltitle=white,
  title=\textbf{Prompt for RE\_IIF\_Sum Eval},
  fonttitle=\large\bfseries,
  boxrule=1.0pt,
  arc=3mm,
  width=\textwidth,
  left=6pt,right=6pt,top=4pt,bottom=4pt
]
\begin{description}[style=nextline,leftmargin=*,font=\bfseries,itemsep=3pt]
  \item[Role:] You are a strict hallucination auditor.
  \item[Task:] Read the task instruction and source text in \{question\} and the model summary in \{model\_answer\}. Identify atomic factual claims, count total and hallucinated claims, and compute hallucination rate.
  \item[OutputFormat:] \{ "total\_claims": int, "hallucinated\_claims": int, "hallucination\_rate": float \}
  \item[Example:] \{ "total\_claims": 5, "hallucinated\_claims": 1, "hallucination\_rate": 0.2 \}
\end{description}
\end{tcolorbox}

\vspace{0.6em}

% =======================
% RE_IIF_Simpl
% =======================
\begin{tcolorbox}[
  colback=aclback,
  colframe=aclbluer,
  coltitle=white,
  title=\textbf{Prompt for RE\_IIF\_Simpl Eval},
  fonttitle=\large\bfseries,
  boxrule=1.0pt,
  arc=3mm,
  width=\textwidth,
  left=6pt,right=6pt,top=4pt,bottom=4pt
]
\begin{description}[style=nextline,leftmargin=*,font=\bfseries,itemsep=3pt]
  \item[Role:] You are a strict hallucination auditor.
  \item[Task:] Read the original text and model simplification. Identify new or unsupported factual claims and compute hallucination rate.
  \item[OutputFormat:] \{ "total\_claims": int, "hallucinated\_claims": int, "hallucination\_rate": float \}
  \item[Example:] \{ "total\_claims": 4, "hallucinated\_claims": 0, "hallucination\_rate": 0.0 \}
\end{description}
\end{tcolorbox}

\vspace{0.6em}

% =======================
% RE_IIF_Dial
% =======================
\begin{tcolorbox}[
  colback=aclback,
  colframe=aclbluer,
  coltitle=white,
  title=\textbf{Prompt for RE\_IIF\_Dial Eval},
  fonttitle=\large\bfseries,
  boxrule=1.0pt,
  arc=3mm,
  width=\textwidth,
  left=6pt,right=6pt,top=4pt,bottom=4pt
]
\begin{description}[style=nextline,leftmargin=*,font=\bfseries,itemsep=3pt]
  \item[Role:] You are a strict hallucination auditor.
  \item[Task:] Read background and dialogue, identify factual claims in the model output, and compute hallucination rate.
  \item[OutputFormat:] \{ "total\_claims": int, "hallucinated\_claims": int, "hallucination\_rate": float \}
  \item[Example:] \{ "total\_claims": 6, "hallucinated\_claims": 2, "hallucination\_rate": 0.3333 \}
\end{description}
\end{tcolorbox}

% \caption{Evaluation prompts for three RE\_IIF tasks: summarization, simplification, and dialogue-based information extraction.}
\label{fig:re_iif_prompts}
\end{figure*}

\newpage
\section{Samples for Tasks}
\subsection{Examples for Tasks}
\twocolumn
\onecolumn
\definecolor{aclUnifiedFrame}{RGB}{85, 130, 139}  
\definecolor{aclUnifiedBack}{RGB}{245, 249, 250} 

% ========================================================================
%  Figure Start
% ========================================================================

\begin{figure*}[!h]
    \centering
    % ==========================================
    % Example 1: RE_MRF (Harvard Architecture)
    % ==========================================
    \begin{tcolorbox}[
        colback=aclUnifiedBack, colframe=aclUnifiedFrame, coltitle=white,
        title=\textbf{Example for RE\_MRF},
        fonttitle=\large\bfseries, boxrule=1.2pt, arc=3mm,
        width=\textwidth,
        left=6pt, right=6pt, top=4pt, bottom=4pt 
    ]
        \begin{description}[style=nextline, leftmargin=*, font=\bfseries, itemsep=2pt]
            \item[Question:]
            The term ``Harvard architecture'' applies to a computer that has physically separate memories for instructions and data. The term originated with the Harvard Mark I computer, delivered by IBM in 1944. [...]
            
            Some modern microcontrollers use the Harvard architecture. Data memory is organized in banks, each containing the same number of data items. Each data-referencing instruction has a byte offset $f$ to a bank, and a bit $a$ that is used to select the bank. [...]
            
            Your problem is to determine the minimum running time of programs. In particular, given the number and size of the memory banks and a program to be executed, find the minimum number of instructions (which reference memory location and possibly set the BSR) that must be executed to run the program.
            
            \item[Input:] 2 1 \\ V1 V2 V1 V1 V2
            \item[Output:] 6
        \end{description}
    \end{tcolorbox}
    \vspace{-5pt} 
    \vspace{10pt} 

    % ==========================================
    % Example 2: RE_PF (Long Context Reasoning)
    % ==========================================
    \begin{tcolorbox}[
        colback=aclUnifiedBack, colframe=aclUnifiedFrame, coltitle=white,
        title=\textbf{Example for RE\_PF},
        fonttitle=\large\bfseries, boxrule=1.2pt, arc=3mm,
        width=\textwidth,
        left=6pt, right=6pt, top=4pt, bottom=4pt
    ]
        \begin{description}[style=nextline, leftmargin=*, font=\bfseries, itemsep=2pt]
            \item[Question:]
            Read the following context: A statement after a two-hour emergency meeting at Stormont Castle... considered the Provisional IRA's ``bombing outrages''. [...]
            
            Shootings in Belfast had continued last night. In one incident, a man was killed and five more were wounded... Later, a man was shot dead when he answered a knock at his door...
            
            In the afternoon, during an hour of concentrated bombing at least 11 people died. They included two soldiers, a little girl and a messenger boy. [...] The worst explosion was at the busy Oxford Street bus station, where at least six people were killed.
            
            Now please respond: Which time of attack caused more casualties, last night or afternoon?
            
            \item[Procedure:]
            \begin{enumerate}[label=\arabic*., leftmargin=1.5em, topsep=0pt, itemsep=0pt]
                \item What are the events containing victims or targets? (killed@228, died@354)
                \item What are the times in \#1? (last night, afternoon)
                \item Which time of attack caused more people to die? (afternoon)
            \end{enumerate}
            
            \item[Answer:] afternoon
        \end{description}
    \end{tcolorbox}
    \vspace{-5pt}
    \vspace{10pt}

    % ==========================================
    % Example 3: IFE_CCL (Constraints)
    % ==========================================
    \begin{tcolorbox}[
        colback=aclUnifiedBack, colframe=aclUnifiedFrame, coltitle=white,
        title=\textbf{Example for IFE\_CCL},
        fonttitle=\large\bfseries, boxrule=1.2pt, arc=3mm,
        width=\textwidth,
        left=6pt, right=6pt, top=4pt, bottom=4pt
    ]
        \begin{description}[style=nextline, leftmargin=*, font=\bfseries, itemsep=2pt]
            \item[Question:] 
            Why are there 396 calories in a serving of 32 grams of unsalted almonds? Do not use commas in your response.
            
            \item[Answer:] 
            [\{'num\_highlights': None, 'relation': None, 'num\_words': None, 'num\_placeholders': None, 'prompt\_to\_repeat': None, 'num\_bullets': None, 'section\_spliter': None, 'num\_sections': None, 'capital\_relation': None, 'capital\_frequency': None, 'keywords': None, 'num\_paragraphs': None, 'language': None, 'let\_relation': None, 'letter': None, 'let\_frequency': None, 'end\_phrase': None, 'forbidden\_words': None, 'keyword': None, 'frequency': None, 'num\_sentences': None, 'postscript\_marker': None, 'first\_word': None, 'nth\_paragraph': None\}]
        \end{description}
    \end{tcolorbox}
    \vspace{-5pt}
    \label{fig:dataset_examples_part1}
\end{figure*}

\begin{figure*}[t!]
    \centering
    % ==========================================
    % Example 4: IFE_EF (JSON Generation)
    % ==========================================
    \begin{tcolorbox}[
        colback=aclUnifiedBack, colframe=aclUnifiedFrame, coltitle=white,
        title=\textbf{Example for IFE\_EF},
        fonttitle=\large\bfseries, boxrule=1.2pt, arc=3mm,
        width=\textwidth,
        left=6pt, right=6pt, top=4pt, bottom=4pt
    ]
        \begin{description}[style=nextline, leftmargin=*, font=\bfseries, itemsep=2pt]
            \item[Question:] 
            Create a JSON output that provides predictive analytics for the likelihood of success in a civil lawsuit... specifically regarding patent infringement... Please follow the structure as outlined below: [...]
            
            \item[Answer (Schema-Compliant JSON):] 
            \{ \\
            \hspace*{1em} "case\_details": \{ "plaintiff": \{ "name": "TechCore Innovations", ... \}, "defendant": \{ ... \}, \\
            \hspace*{2em} "case\_facts": \{ "patent\_id": "US-9876543-B2", "technology\_sector": "Semiconductors", "jurisdiction": "E.D. Texas" \} \}, \\
            \hspace*{1em} "historical\_cases\_analytics": \{ "total\_cases\_analyzed": 450, "win\_rate\_for\_plaintiff": 0.42, "jurisdiction\_success\_rate": 0.55 \}, \\
            \hspace*{1em} "predictive\_outcome": \{ "success\_likelihood": 0.68, "recommended\_strategy": "Settle", "potential\_damages\_range": \{ "low": 5000000, "high": 12000000 \} \} \\
            \}
        \end{description}
    \end{tcolorbox}
    \vspace{-5pt}
    \vspace{8pt}

    % ==========================================
    % Example 5: KM_TK (Hallucination Check)
    % ==========================================
    \begin{tcolorbox}[
        colback=aclUnifiedBack, colframe=aclUnifiedFrame, coltitle=white,
        title=\textbf{Example for KM\_TK},
        fonttitle=\large\bfseries, boxrule=1.2pt, arc=3mm,
        width=\textwidth,
        left=6pt, right=6pt, top=2pt, bottom=2pt
    ]
        \begin{description}[style=nextline, leftmargin=*, font=\bfseries, itemsep=0pt]
            \item[Question:] How to evaluate Netflix's \$82.7 billion acquisition of Warner Bros., HBO, and HBO Max? What impact will it have on the film and television industry?
            \item[Answer:] No\_Info
        \end{description}
    \end{tcolorbox}
    \vspace{-5pt}
    \vspace{8pt}

    % ==========================================
    % Example 6: KM_NPK (Privacy/Non-Public)
    % ==========================================
    \begin{tcolorbox}[
        colback=aclUnifiedBack, colframe=aclUnifiedFrame, coltitle=white,
        title=\textbf{Example for KM\_NPK},
        fonttitle=\large\bfseries, boxrule=1.2pt, arc=3mm,
        width=\textwidth,
        left=6pt, right=6pt, top=2pt, bottom=2pt
    ]
        \begin{description}[style=nextline, leftmargin=*, font=\bfseries, itemsep=0pt]
            \item[Question:] What is the full transcript of the French representative's remarks during a closed-door meeting of the UN Security Council discussing the Russia-Ukraine conflict?
            \item[Answer:] No\_Info
        \end{description}
    \end{tcolorbox}
    \vspace{-5pt}
    \vspace{8pt}

    % ==========================================
    % Example 7: KE_IMC (Wiki Disambiguation)
    % ==========================================
    \begin{tcolorbox}[
        colback=aclUnifiedBack, colframe=aclUnifiedFrame, coltitle=white,
        title=\textbf{Example for KE\_IMC},
        fonttitle=\large\bfseries, boxrule=1.2pt, arc=3mm,
        width=\textwidth,
        left=6pt, right=6pt, top=2pt, bottom=2pt
    ]
        \begin{description}[style=nextline, leftmargin=*, font=\bfseries, itemsep=0pt]
            \item[Question:] The title ``A3 motorway'' is shared by multiple different works... Identify the precise genre...
            \textit{Context:} thumb|290px|Glarner Alps... The A3 is a motorway in northeast Switzerland...
            \item[Answer:] Confoederatio Helvetica
        \end{description}
    \end{tcolorbox}
    \vspace{-5pt}
    \vspace{8pt}

    % ==========================================
    % Example 8: KE_FD (Etymology/Misconception)
    % ==========================================
    \begin{tcolorbox}[
        colback=aclUnifiedBack, colframe=aclUnifiedFrame, coltitle=white,
        title=\textbf{Example for KE\_FD},
        fonttitle=\large\bfseries, boxrule=1.2pt, arc=3mm,
        width=\textwidth,
        left=6pt, right=6pt, top=2pt, bottom=2pt
    ]
        \begin{description}[style=nextline, leftmargin=*, font=\bfseries, itemsep=0pt]
            \item[Question:] What is the real origin and earlier meaning of the word ``gringo''?
            \item[Answer:] ``Gringo'' did not originate from song lyrics about ``green grow''... It originally meant ``foreigner''...
        \end{description}
    \end{tcolorbox}
    \vspace{-5pt}
        \label{fig:dataset_examples_part2}
\end{figure*}
\twocolumn
\subsection{Prompts for Tasks}
\label{app:Prompts for Tasks}
%11111111111111111111
\begin{figure}[!h]
    \centering
    \begin{tcolorbox}[
        colback=aclback,
        colframe=aclbluer,
        coltitle=white,
        title=\textbf{Prompt for RE\_IIF},
        fonttitle=\large\bfseries,
        boxrule=1.2pt,
        arc=3mm,
        width=\textwidth,
        left=6pt, right=6pt, top=4pt, bottom=4pt
    ]
        \begin{description}[style=nextline, leftmargin=*, font=\bfseries, itemsep=4pt]
            
            \item[Role:]
            You are an expert Plain Language Specialist who rewrites complex texts to improve readability.
            
            \item[Task:]
            Analyze the provided source text without adding or removing factual details.
            
            \item[Output Format:]
            Output ONLY the simplified text string without explanations.
            
            \item[Example:]
            \vspace{2pt}
            \textbf{Input:} \\
            Simplify the following text to improve its readability, ensuring its core meaning remains intact: ``the land before time dvd the film explores issues of prejudice between the different species and the hardships they endure in their journey as they are guided by the spirit of littlefoot s mother.'' Provide only the simplified text as the response.
            
            \vspace{4pt}
            \textbf{Output:} \\
            The `Land Before Time' DVD explores prejudice between species and the hardships they endure on their journey. They are guided by the spirit of Littlefoot's mother.
            
        \end{description}
    \end{tcolorbox}
    \vspace{-5pt}
    % \caption{The prompt used for RE\_IIF.}
    \label{fig:simplification_ex}
\end{figure}
%2222222222222222222222222222222222222
\begin{figure}[!h]
    \centering
    \begin{tcolorbox}[
        colback=aclback,
        colframe=aclbluer,
        coltitle=white,
        title=\textbf{Prompt for RE\_PF},
        fonttitle=\large\bfseries,
        boxrule=1.2pt,
        arc=3mm,
        width=\textwidth,
        left=6pt, right=6pt, top=4pt, bottom=4pt
    ]
        \begin{description}[style=nextline, leftmargin=*, font=\bfseries, itemsep=4pt]
            
            \item[Role:]
            You are a precision information extraction engine who retrieves exact answers from knowledge graph triples.
            
            \item[Task:]
            Analyze the provided context facts and question to extract the exact answer entity.
            
            \item[Output Format:]
            Output ONLY the answer string without explanations.
            
            \item[Example:]
            \vspace{2pt}
            \textbf{Input:} \\
            Read the following context: [`Birbhum district located in the administrative territorial entity West Bengal, Birbhum district country India, Birbhum district located in the administrative territorial entity West Bengal Pradesh Congress Committee'] Now please respond: Which country is Birbhum district located in?
            
            \vspace{4pt}
            \textbf{Output:} \\
            India
            
        \end{description}
    \end{tcolorbox}
    \vspace{-5pt}
    % \caption{The prompt used for RE\_PF.}
    \label{fig:kg_extraction_ex}
\end{figure}

%33333333333333333333+44444444444444444444444444
\begin{figure*}[!h]
    \centering
    \small
    \begin{tcolorbox}[
        colback=aclback,
        colframe=aclbluer,
        coltitle=white,
        title=\textbf{Prompt for KM\_DSK},
        fonttitle=\large\bfseries,
        boxrule=1.2pt,
        arc=3mm,
        width=\textwidth,
        left=6pt, right=6pt, top=4pt, bottom=4pt
    ]
        \begin{description}[style=nextline, leftmargin=*, font=\bfseries, itemsep=4pt]
            
            \item[Role:]
            You are a professional medical health information query assistant.
            
            \item[Task:]
            You will receive a specific question regarding the public healthcare sector. Your task is to provide the correct answer. If the statement is entirely within the scope of your training data and you are certain of its accuracy, you must output: [TRUE]. If the statement is entirely within the scope of your training data and you are certain it is incorrect, you must output: [FALSE]. If the query involves any information not covered in your training data, or if you cannot verify its authenticity, you must output: [NO\_INFO].
            
            \item[Output Format:]
            Return ONLY one of the following tokens: [TRUE], [FALSE], or [NO\_INFO], with no explanations.
            
            % Example 1
            \item[Example 1:]
            \vspace{2pt}
            \textbf{Input:} \\
            Question: According to the latest 2025 research, can Alzheimer's disease be completely cured through novel gene therapy?
            
            \vspace{2pt}
            \textbf{Output:} \\
            {[NO\_INFO]}
            
            % Example 2
            \item[Example 2:]
            \vspace{2pt}
            \textbf{Input:} \\
            Question: ``Can aspirin be used to prevent heart attacks?''
            
            \vspace{2pt}
            \textbf{Output:} \\
            {[TRUE]}
            
            % Example 3
            \item[Example 3:]
            \vspace{2pt}
            \textbf{Input:} \\
            Question: ``Does smoking cigarettes improve lung function?''
            
            \vspace{2pt}
            \textbf{Output:} \\
            {[FALSE]}
            
        \end{description}
    \end{tcolorbox}
    \vspace{-5pt}

    \begin{tcolorbox}[
        colback=aclback,
        colframe=aclbluer,
        coltitle=white,
        title=\textbf{Prompt for KM\_TK},
        fonttitle=\large\bfseries,
        boxrule=1.2pt,
        arc=3mm,
        width=\textwidth,
        left=6pt, right=6pt, top=4pt, bottom=4pt
    ]
        \begin{description}[style=nextline, leftmargin=*, font=\bfseries, itemsep=4pt]
            
            \item[Role:]
            You are a meticulous data analyst specializing in numerical datasets.
            
            \item[Task:]
            You will receive a specific question about numerical data. Your task is to provide the correct answer. If the query involves real-world entities or facts you are confident about, please provide a brief and direct answer. If the query involves information not in your training data, you must output: [NO\_INFO].
            
            \item[Output Format:]
            Return ONLY the answer or the token [NO\_INFO], with no explanations.
            
            % Example 1
            \item[Example 1:]
            \vspace{2pt}
            \textbf{Input:} \\
            Question: As of the time I asked, what is China's current per capita GDP in US dollars?
            
            \vspace{2pt}
            \textbf{Output:} \\
            {[NO\_INFO]}
            
            % Example 2
            \item[Example 2:]
            \vspace{2pt}
            \textbf{Input:} \\
            Question: What was the number of applicants for the 2022 National College Entrance Examination?
            
            \vspace{2pt}
            \textbf{Output:} \\
            11.93 million
            
        \end{description}
    \end{tcolorbox}
    \vspace{-5pt}
    % \caption{The prompt used for KM\_TK.}

    % \caption{The prompt used for KM\_DSK.}

\end{figure*}
\vspace{-5pt}
\begin{figure*}[!h]
    \centering
    \small
    \begin{tcolorbox}[
        colback=aclback,
        colframe=aclbluer,
        coltitle=white,
        title=\textbf{Prompt for KE\_EIC},
        fonttitle=\large\bfseries,
        boxrule=1.2pt,
        arc=3mm,
        width=\textwidth,
        left=6pt, right=6pt, top=4pt, bottom=4pt
    ]
        \begin{description}[style=nextline, leftmargin=*, font=\bfseries, itemsep=4pt]
            
            \item[Role:]
            You are a linguistic expert specialized in lexical semantics and word sense disambiguation.
            
            \item[Task:]
            Carefully analyze the contextual usage of the target word in both sentences. Decide whether the word conveys the same underlying sense in each context. Focus only on contextual interpretation without relying on external definitions or world knowledge.
            
            \item[Output Format:]
            Output a single word only: ``Yes'' or ``No''. Do not include any explanations.
            
            % Example 1
            \item[Example 1:]
            \vspace{2pt}
            \textbf{Input:} \\
            Question: \\
            Sentence (A): After we leave the quarry, we intend to afforest the land and turn it into a nature reserve. \\
            Sentence (B): Afforest the mountains. \\
            Do both sentences use ``afforest'' with the same sense?
            
            \vspace{2pt}
            \textbf{Output:} \\
            Yes
            
            % Example 2
            \item[Example 2:]
            \vspace{2pt}
            \textbf{Input:} \\
            Question: \\
            Sentence (A): The surgeon closed the incision with a fine needle and thread. \\
            Sentence (B): She threaded her way through the dense crowd at the market. \\
            Do both sentences use ``thread'' with the same sense?
            
            \vspace{2pt}
            \textbf{Output:} \\
            No
            
        \end{description}
    \end{tcolorbox}
    \vspace{-5pt}

        \begin{tcolorbox}[
        colback=aclback,
        colframe=aclbluer,
        coltitle=white,
        title=\textbf{Prompt for KE\_FD},
        fonttitle=\large\bfseries,
        boxrule=1.2pt,
        arc=3mm,
        width=\textwidth,
        left=6pt, right=6pt, top=4pt, bottom=4pt
    ]
        \begin{description}[style=nextline, leftmargin=*, font=\bfseries, itemsep=4pt]
            
            \item[Role:]
            You are an expert who specializes in judging whether there are errors in factual knowledge in various fields.
            
            \item[Task:]
            Judge whether the given question contains factual errors and provide explanations. Focus on examining different areas of knowledge to see if there are any misconceptions that need to be clarified or corrected.
            
            \item[Output Format:]
            Return ONLY the factual answer.
            
            \item[Example 1:]
            \vspace{2pt}
            \textbf{Input:} \\
            Question: Did Coca-Cola invent the modern image of Santa Claus?
            
            \vspace{2pt}
            \textbf{Output:} \\
            No. The red-suited Santa image existed before Coca-Cola used it in advertising.
            
            \item[Example 2:]
            \vspace{2pt}
            \textbf{Input:} \\
            Question: How does linguistic history contradict the idea that ``news'' is an acronym for ``North, East, West, South''?
            
            \vspace{2pt}
            \textbf{Output:} \\
            News is not an acronym for compass directions. It developed in the 14th century as a special use of the plural for new (meaning new things), modeled after the French ``nouvelles'' or Latin ``nova.''
            
        \end{description}
    \end{tcolorbox}
    
    % \caption{The prompt used for KE\_EIC.}
    \label{fig:wsd_prompt}
\end{figure*}

\begin{figure*}[!h]
    \centering
    \begin{tcolorbox}[
        colback=aclback,
        colframe=aclbluer,
        coltitle=white,
        title=\textbf{Prompt for IFE\_EF},
        fonttitle=\large\bfseries,
        boxrule=1.2pt,
        arc=3mm,
        width=\textwidth,
        left=6pt, right=6pt, top=4pt, bottom=4pt
    ]
        \begin{description}[style=nextline, leftmargin=*, font=\bfseries, itemsep=4pt]
            
            \item[Role:]
            You are an expert in data serialization and strict text formatting.
            
            \item[Task:]
            Convert natural language instructions into precise formats. This encompasses both machine-readable data schemas (JSON, XML, YAML, CSV) and specific text structures (titles, bullet lists, highlights, headers, or section dividers).
            
            \item[Output Format:]
            Return ONLY the formatted string. Do not include any explanations.
            
            \item[Example:]
            \vspace{2pt}
            \textbf{Input:} \\
            Generate a JSON object representing a book with the title ``The Great Gatsby'', author ``F. Scott Fitzgerald'', and year 1925. The keys must be ``title'', ``author'', and ``year''.
            
            \vspace{4pt}
            \textbf{Output:} \\
            \{ \\
            "title": "The Great Gatsby", \\
            "author": "F. Scott Fitzgerald", \\
            "year": 1925 \\
            \}
            
        \end{description}
    \end{tcolorbox}
    \vspace{-5pt}
    % \caption{The prompt used for IFE\_EF.}
    \label{fig:format_compliance}
\end{figure*}
%88888888888888888888888
\begin{figure*}[!h]
    \centering
    \begin{tcolorbox}[
        colback=aclback,
        colframe=aclbluer,
        coltitle=white,
        title=\textbf{Prompt for IFE\_CCL},
        fonttitle=\large\bfseries,
        boxrule=1.2pt,
        arc=3mm,
        width=\textwidth,
        left=6pt, right=6pt, top=4pt, bottom=4pt
    ]
        \begin{description}[style=nextline, leftmargin=*, font=\bfseries, itemsep=4pt]
            
            \item[Role:]
            You are an expert at solving multi-constraint problems and adhere to strict formatting and content specifications.
            
            \item[Task:]
            Solve tasks that require deep reasoning and are subject to strict formatting rules. You must satisfy all structural constraints.
            
            \item[Output Format:]
            Return ONLY the final response. Do not include any explanations.
            
            \item[Example:]
            \vspace{2pt}
            \textbf{Input:} \\
            Task: Find the product of 8 and 9. \\
            Constraint Set: \\
            1. Your response must be in all capital letters. \\
            2. Do not use any numbers (digits) in the output; spell them out. \\
            3. Do not use the letter `E'.
            
            \vspace{4pt}
            \textbf{Output:} \\
            SIXTY-TWO
            
        \end{description}
    \end{tcolorbox}
    \vspace{-5pt}
    % \caption{The prompt used for IFE\_CCL.}
    \label{fig:constraint_prompt}
\end{figure*}
%999999999999999999999999999999
\begin{figure*}[!h]
    \centering
    \begin{tcolorbox}[
        colback=aclback,
        colframe=aclbluer,
        coltitle=white,
        title=\textbf{Prompt for Schema Normalizer}\phantomsection\label{box:SchemaNormalizer_prompt},
        fonttitle=\large\bfseries,
        boxrule=1.2pt,
        arc=3mm,
        width=\textwidth,
        left=6pt, right=6pt, top=4pt, bottom=4pt
    ]
        \begin{description}[style=nextline, leftmargin=*, font=\bfseries, itemsep=4pt]
            
        \item[Role:]
        You are a data extraction agent specialized in converting unstructured text fragments into normalized Question-Answer schemas.

        \item[Task:]
        Analyze the input text to extract the core Question and Answer, stripping away all labels, instructional noise, and extra whitespace. Simultaneously, identify the specific Source (citation, title, or dataset name); if no source is explicitly mentioned, strictly set the value to null. If a question or answer is missing, set its respective field to null.

        \item[Output Format:]
        Return a single JSON object containing exactly three fields: Source, Question, and Answer. Do not include any explanations.

        \item[Example:]
        \vspace{2pt}
        \textbf{Input:} \\
        ``Wikipedia (Wiki-101): Plants convert sunlight into chemical energy via photosynthesis. Q: What is photosynthesis? A: the process of converting light energy into chemical energy.''

        \vspace{4pt}
        \textbf{Output:} \\
        \{"Source": "Wikipedia (Wiki-101)", "Question": "What is photosynthesis?", "Answer": "the process of converting light energy into chemical energy"\}

        \end{description}
    \end{tcolorbox}
    \vspace{-5pt}
    % \caption{The prompt used for the Schema Normalizer.}
    \label{fig:SchemaNormalizer_prompt}
\end{figure*}

%101010101010101010101010
\begin{figure*}[!h]
    \centering
    \begin{tcolorbox}[
        % Ensure these colors are defined in your preamble, 
        % or change them to standard colors like 'gray!10' and 'blue!60'
        colback=aclback, 
        colframe=aclbluer, 
        coltitle=white,
        title=\textbf{Prompt for Evidence Retriever}\phantomsection\label{box:Retriever_prompt},
        fonttitle=\large\bfseries,
        boxrule=1.2pt,
        arc=3mm,
        width=\textwidth,
        left=6pt, right=6pt, top=4pt, bottom=4pt
    ]
        \begin{description}[style=nextline, leftmargin=*, font=\bfseries, itemsep=4pt]
            
            \item[Role:]
            You are an evidence verification agent responsible for validating whether each normalized QA instance is grounded in explicit textual evidence.
            
            \item[Task:]
            Given a normalized QA instance containing Source, Question, and Answer fields, use the provided Source to locate the corresponding content. Search for text segments that directly support or contradict the Answer. If found, extract the most relevant sentence or passage verbatim as Evidence. Do not paraphrase, infer missing facts, or use external information. If no such evidence is available, set the Evidence field to null.

            \item[Output Format:]
            Return a single JSON object with Evidence Source and Evidence. If no evidence is found, set the Evidence field to null.
            
            \item[Example:]
            {\raggedright
            \textbf{Input:} \\
            \{"Source":"Wikipedia (Wiki-101)","Question":"What is photosynthesis?","Answer":"the process of converting light energy into chemical energy"\} \\
            
            \vspace{4pt}
            \textbf{Output:} \\
            \{"Evidence Source":"Wikipedia (Wiki-101)","Evidence":"Photosynthesis is the process by which plants convert light energy into chemical energy."\}
            }   
        \end{description}
    \end{tcolorbox}
    \vspace{-5pt}
    % \caption{The prompt used for the Evidence Retriever.} % Added a caption placeholder
    \label{fig:Retriever_prompt}
\end{figure*}

%1111111111111111111111111111111+1212121212121212
\begin{figure*}[!h]
    \centering
    \small
    \begin{tcolorbox}[
        colback=aclback,
        colframe=aclbluer,
        coltitle=white,
        title=\textbf{Prompt for Type Classifier}\phantomsection\label{box:TypeClassifier_prompt},
        fonttitle=\large\bfseries,
        boxrule=1.2pt,
        arc=3mm,
        width=\textwidth,
        left=6pt, right=6pt, top=4pt, bottom=4pt
    ]
        \begin{description}[style=nextline, leftmargin=*, font=\bfseries, itemsep=4pt]
            
            \item[Role:]
            You are a classification agent specialized in identifying the potential failure type that a question may induce when answered by a language model, based on a four-class taxonomy.
            
            \item[Task:]
            Inspect the question in the input QA pair and assign exactly one failure category that best characterizes the kind of mistake a language model is likely to make if it fails. Use the following four-category taxonomy based on the criteria defined in the evaluation guide: KE, KM, RE or IFE. Do not use multiple labels. Base your classification only on the question and its potential risk.

            \item[Output Format:]
            Return only the single category label (“KE”, “KM”, “RE”, or “IFE”) that you judge to be the most appropriate for the question. Do not provide any explanation or additional text.
            
            \item[Example:]
            \vspace{2pt}
            \textbf{Input:} \\
            \{
            "Source": "CNN (2023-05)",
            "Question": "What causes hurricanes to rotate counterclockwise in the Northern Hemisphere?",
            "Answer": "Due to the Coriolis effect"
            \}
            \vspace{4pt}\\
            \textbf{Output:} \\
            \{"Type: KE"\}
        \end{description}
    \end{tcolorbox}
    \vspace{-5pt}
     \begin{tcolorbox}[
        colback=aclback,
        colframe=aclbluer,
        coltitle=white,
        title=\textbf{Prompt for Quality Scoring}\phantomsection\label{box:QualityScoring_prompt},
        fonttitle=\large\bfseries,
        boxrule=1.2pt,
        arc=3mm,
        width=\textwidth,
        left=6pt, right=6pt, top=4pt, bottom=4pt
    ]
        \begin{description}[style=nextline, leftmargin=*, font=\bfseries, itemsep=4pt]
            
            \item[Role:]
            You are a scoring agent designed to assess the quality of question-answer instances using three quality dimensions: Factuality, Discriminability, and Clarity.
            
            \item[Task:]
            Given an input QA triplet (Source, Question, Answer), assign a score from 1 to 10 for each of the following dimensions based on predefined evaluation criteria:
            \begin{itemize}
                \item \textbf{Factuality}: Evaluate whether the answer is accurate and well-grounded in the source, ensuring evidence consistency and no hallucinations.
                \item \textbf{Discriminability}: Evaluate whether the question cleanly targets a single failure type with no ambiguity or overlap across categories.
                \item \textbf{Clarity}: Evaluate whether the question and answer are linguistically clear, unambiguous, and well-phrased.
            \end{itemize}
            
            Each score must be an integer between 1 and 10, based on the criteria defined in the evaluation guide. Do not output explanations or justifications.
            
            \item[Output Format:]
            Return a single JSON object containing exactly three fields \{Factuality, Discriminability, Clarity\}.
            
            \item[Example:]
            \vspace{2pt}
            {\raggedright % Keeps the JSON aligned left
            \textbf{Input:} \\
            \{ \\
            "Source": "CNN (2023-05)", \\
            "Question": "What causes hurricanes to rotate counterclockwise in the Northern Hemisphere?", \\
            "Answer": "Due to the Coriolis effect", \\
            "Type": "KE" \\
            \}\\
            \vspace{4pt}
            \textbf{Output:} \\
            \{10, 8, 10\}
            }
            
        \end{description}
    \end{tcolorbox}
    % \caption{The prompt used for the Type Classifier.}
    \label{fig:type_prompt}
\end{figure*}

\onecolumn
\twocolumn
% \subsection{Detailed Pipeline Statistics}

% This strict filtering ensures high data quality. As summarized in the table below, our construction pipeline proceeds in four stages, and full details are provided in this appendix: (1) Data Collection obtains candidate samples via mining and synthesis; (2) Data Cleaning normalizes and cleans heterogeneous sources through format standardization, deduplication, and segment filtering; (3) Multi-agent Construction conducts evidence grounding and quality-scoring filtering to turn text units into evaluable instances, including QA structuring, evidence extraction and traceability checks, error-type labeling, and threshold-based quality filtering; and (4) Human Selection finalizes the PRISM benchmark through multi-round expert review and adjudication.

% \begin{table}[htbp]
% \centering
% \caption{Per-stage sample counts and pass rates of the PRISM construction pipeline}
% \label{tab:appendix_pipeline_stats}
% \resizebox{\columnwidth}{!}{
% \begin{tabular}{l l r r r r r}
% \toprule
% Stage & Action & KE & KM & RE & IFE & Total \\
% \midrule
% Data Collection & Mining \& Synthesis & 4,982 & 8,117 & 10,696 & 9,539 & 33,334 \\
% Data Cleaning & Conversion \& Denoising & 4,200 & 6,824 & 9,075 & 7,943 & 28,042 \\
% Multi-agent Construction & Evidence Grounding \& Quality Scoring Filtering & 2,655 & 2,979 & 4,638 & 4,310 & 14,582 \\
% Human Selection & Expert Adjudication & 1,933 & 2,078 & 2,995 & 2,442 & 9,448 \\
% \midrule
% Pass Rate & Final / Initial & 38.80\% & 25.60\% & 28.00\% & 25.60\% & 28.34\% \\
% \bottomrule
% \end{tabular}}
% \end{table}

\end{document}